\documentclass[aps,prl,reprint,superscriptaddress]{revtex4-2}

\usepackage{scalerel}
\usepackage{graphicx}
\usepackage{dcolumn}
\usepackage{bm}
\usepackage{color}
\usepackage{xcolor}
\usepackage{epstopdf}
\usepackage{gensymb}
\usepackage{ulem}
\usepackage{float}
\usepackage{xr}
\usepackage{sidecap}
\usepackage{mathrsfs}
\usepackage{amsmath}
\usepackage{amssymb}
\usepackage{amsthm}
\usepackage{enumerate}
\usepackage{soul}
\usepackage{booktabs, array}
\usepackage{multirow}
\usepackage{mathtools, bm, nicefrac, dsfont}
\usepackage{siunitx}
\usepackage{afterpage}
\usepackage{tikz}
\usetikzlibrary{automata,positioning,shapes,decorations}
\usepackage{chemfig}
\DeclareSIUnit{\tss}{\text{timesteps}}
\usepackage{bigstrut}
\begin{document}


\title{The ``Pac-Man'' Gripper: Tactile Sensing and Grasping through Thin-Shell Buckling}

\author{Kieran Barvenik}
     \affiliation{%
 Department of Mechanical Engineering,
    University of Maryland, College Park, MD 20742, USA.
}
\author{Zachary Coogan}
     \affiliation{%
 Department of Mechanical Engineering, University of Maryland, College Park, MD 20742, USA.
}

\author{Gabriele Librandi}
     \affiliation{%
 Independent Researcher.
}

\author{Matteo Pezzulla}
     \affiliation{%
Department of Mechanical and Production Engineering,    Aarhus University, Denmark.
}

\author{Eleonora Tubaldi}
     \email{etubaldi@umd.edu}
 \affiliation{%
 Department of Mechanical Engineering, University of Maryland, College Park, MD 20742, USA.
}
      \affiliation{%
Maryland Robotics Center, University of Maryland, College Park, MD 20742, USA.\\
 }
  \affiliation{%
Department of Electrical Engineering, University of Maryland, College Park, MD 20742, USA.\\
}
 \affiliation{%
Robert E. Fischell Institute of Biomedical Devices, University of Maryland, College Park, MD 20742, USA.}

\begin{abstract}
    Soft and lightweight grippers have greatly enhanced the performance of robotic manipulators in handling complex objects with varying shape, texture, and stiffness.
However, the combination of universal grasping with passive sensing capabilities still presents challenges.
To overcome this limitation, we introduce a fluidic soft gripper, named the ``Pac-Man'' gripper, based on the buckling of soft, thin hemispherical shells.
Leveraging a single fluidic pressure input, the soft gripper can encapsulate slippery and delicate objects while passively providing information on this physical interaction. Guided by analytical, numerical, and experimental tools, we explore the novel grasping principle of this mechanics-based soft gripper. First, we characterize the buckling behavior of a free hemisphere as a function of its geometric parameters. Inspired by the free hemisphere's two-lobe mode shape ideal for grasping purposes, we demonstrate that the gripper can perform dexterous manipulation and gentle gripping of fragile objects in confined environments. Last, we prove the soft gripper's embedded capability of detecting contact, grasping, and release conditions during the interaction with an unknown object.
This simple buckling-based soft gripper opens new avenues for the design of adaptive gripper morphologies with applications ranging from medical and agricultural robotics to space and underwater exploration.

\end{abstract}

\maketitle

\section{Introduction}
Shape and function in natural organisms converge towards geometries with advantageous mechanical characteristics \cite{thompson_growth_1992,ball_natures_2009,banavar_form_2014}, leading to the formation of structural patterns across the natural world \cite{ball_natures_2009}. 
Inspired by these natural design principles, soft robots transcend the limitations of conventional rigid systems and create new possibilities for diverse functional properties. 
State-of-the art soft grippers have enhanced the functionalities of robotic manipulators through precision grasping \cite{yang_grasping_2021, hong_boundary_2022}, adaptable gripping capabilities \cite{brown_universal_2010, li_vacuum-driven_2019, becker_active_2022}, fragile object manipulation \cite{sinatra_ultragentle_2019, yoder_soft_2023}, and embedded soft sensing elements \cite{truby_soft_2018, li_resonant_2022}. Additionally, soft materials enable the integration of unique properties like self-healing \cite{acome_hydraulically_2018,liu_polyionic_2020,terryn_self-healing_2017}, mechanical computing \cite{yasuda2021mechanical, lin2023recent}, and intrinsic programmability \cite{zhuo_complex_2020,yang_grasping_2021,ze_magnetic_2020,wang_soft_2018} to reimagine the capabilities of robotic end effectors.
Current soft grippers are designed to be actuated with a variety of stimuli, including magnetic and electric fields \cite{zhang_magnetic_2019,ze_magnetic_2020,alapan_reprogrammable_2020}, temperature gradients \cite{cheng_bilayer-type_2018,breger_self-folding_2015}, light energy \cite{pilz_da_cunha_untethered_2019, pan_3d_2021}, and pneumatic or hydraulic inputs \cite{ilievski_soft_2011, li_fluid-driven_2017, deimel_novel_2016}.

While these actuation schemes increase the functional diversity of soft robots, they all present unique challenges ranging from slow response speed and low output force for electric, thermal, and pneumatic inputs to limited range of actuation for magnetically-responsive materials, respectively.
Embedding mechanical instabilities within soft devices can overcome these limitations by triggering large deformations and shape transitions in response to small changes in external stimuli \cite{pal2021exploiting}.
Elements such as prebuckled beams \cite{yang_buckling_2015}, multistable linkages \cite{tang_leveraging_2020,lin_bioinspired_2021}, or snapping doubly curved shells \cite{qi_sea-anemone-inspired_2022,jin_ultrafast_2023} have been used to produce programmable rapid reconfigurations for gripping tasks.
Despite these advancements, current limitations include difficulties in enabling soft robots with multiple functional behaviors, such as fast grasping of delicate objects with a sense of touch in complex environments.

To overcome these limitations, inspiration can be once again drawn from the natural world. 
Among the numerous structural patterns in nature, hemisphere-like geometries appear across scales ranging from the microscale of viruses \cite{lidmar_virus_2003} and bacteria \cite{glaenzer_peldor_2017} to the macroscale of fungi \cite{schaechter2002stroll}, pollen grains \cite{katifori_foldable_2010}, plants \cite{forterre_how_2005}, and sea creatures \cite{nakane_cytoskeletal_2007}.
The folding of such hemispheres and similar ``Pac-Man'' shapes represent largely observed encapsulation mechanisms among proteins and immune cells to capture sugar molecules \cite{glaenzer_peldor_2017} and grab cellular debris and pathogens \cite{tauzin_redox_2014}.
These patterns continue at the macroscale, where hemispherical structures facilitate diverse behaviors, from the attachment of marine leeches \cite{burreson_marine_2020} and the adhesion of \textit{Octopus vulgaris} suckers \cite{tramacere_structure_2014} to the locomotion of jellyfish \cite{costello_hydrodynamics_2021}.

\begin{figure*}[t]
    \centering
    \includegraphics[width=\textwidth,keepaspectratio]{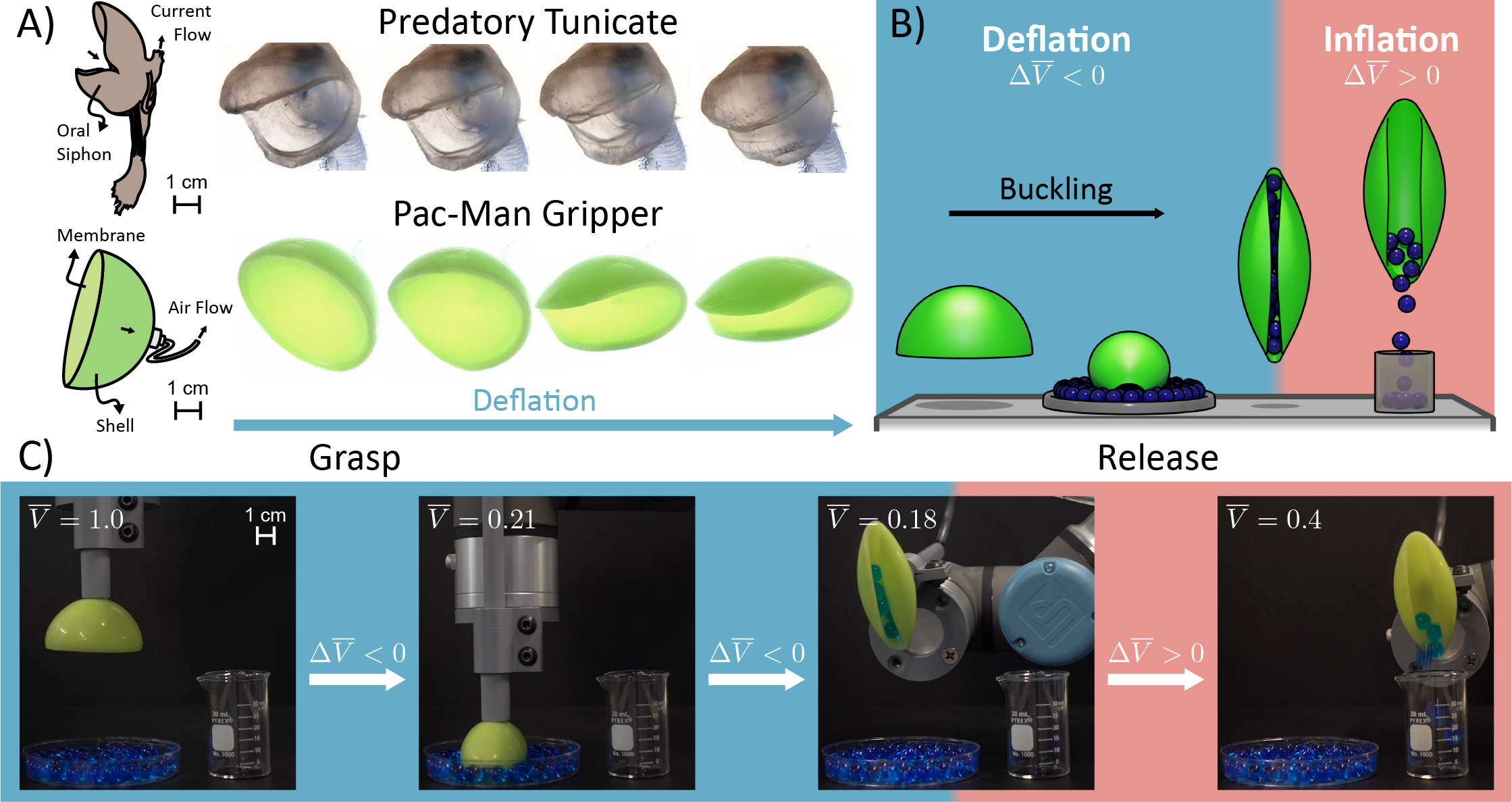}
    \caption{\textbf{Bioinspiration and Function of the ``Pac-Man''  Gripper.} \textbf{A)} Folding shape and feeding behavior of the predatory \textit{Benthic tunicate} represented in a schematic  \cite{gage_deep-sea_1999} and in images from \textit{The Blue Planet} documentary \cite{fothergill_blue_2001}. Analogy with the hemispherical gripper that mimics the shape and function of the deep sea tunicate to gently manipulate objects. \textbf{B)} The buckling-induced grasping mechanism of the hemispherical gripper based on the change in volume $\Delta \overline{V}=\Delta V/V_0$ of the fluidic cavity with initial volume $V_0$. \textbf{C)} Experimental snapshots of the hemispherical gripper grasping, protecting, and releasing wet hydrogel spheres.}
    \label{fig:f1}
\end{figure*}

\textit{Megalodicopia hians}, a variant of the deep-sea \textit{Benthic tunicate}, developed large hemispherical oral siphons to catch floating organic particles (Fig. \ref{fig:f1}A).
Aided by the flow of water, these oral siphons collapse around food, providing an efficient and energetically favorable feeding mechanism \cite{gage_deep-sea_1999}.
Here, inspired by the predatory deep-sea tunicate (\textit{Ascidian, Ascidiacea}), we introduce a new class of soft, pneumatic, buckling-based hemispherical gripper (Fig. \ref{fig:f1}B - C and movie S1), named the ``Pac-Man'' gripper, that can be programmed to selectively grasp, encapsulate, and release fragile and slippery objects in unstructured and confined environments, all while providing electronics-free tactile sensing. 

To explore this new grasping strategy, we first demonstrate with a combination of analytical and numerical models that the free hemisphere experiences a buckling instability at a critical pressure that follows a predictable cubic trend with respect to slenderness ratio.
We then combine numerical and experimental methods to investigate the behavior of the buckling-based soft inflatable gripper by applying a thin membrane on the equatorial plane of the hemisphere.
We identify geometric parameters that enable constructive buckling for functional and universal grasping with a systematic exploration of the design space.
Then, we show how the properties of the soft inflatable gripper can be leveraged for precise gripping and manipulation tasks of delicate objects in complex and confined environments.
Last, we demonstrate that the ``Pac-Man'' gripper showcases an electronics-free sense of touch while it performs dexterous grasping and manipulation tasks.



\section{Results}
\subsection{Buckling Behavior of Free Thin-Shell Domes}
\begin{figure*}
    \centering
    \includegraphics[width=\textwidth,keepaspectratio]{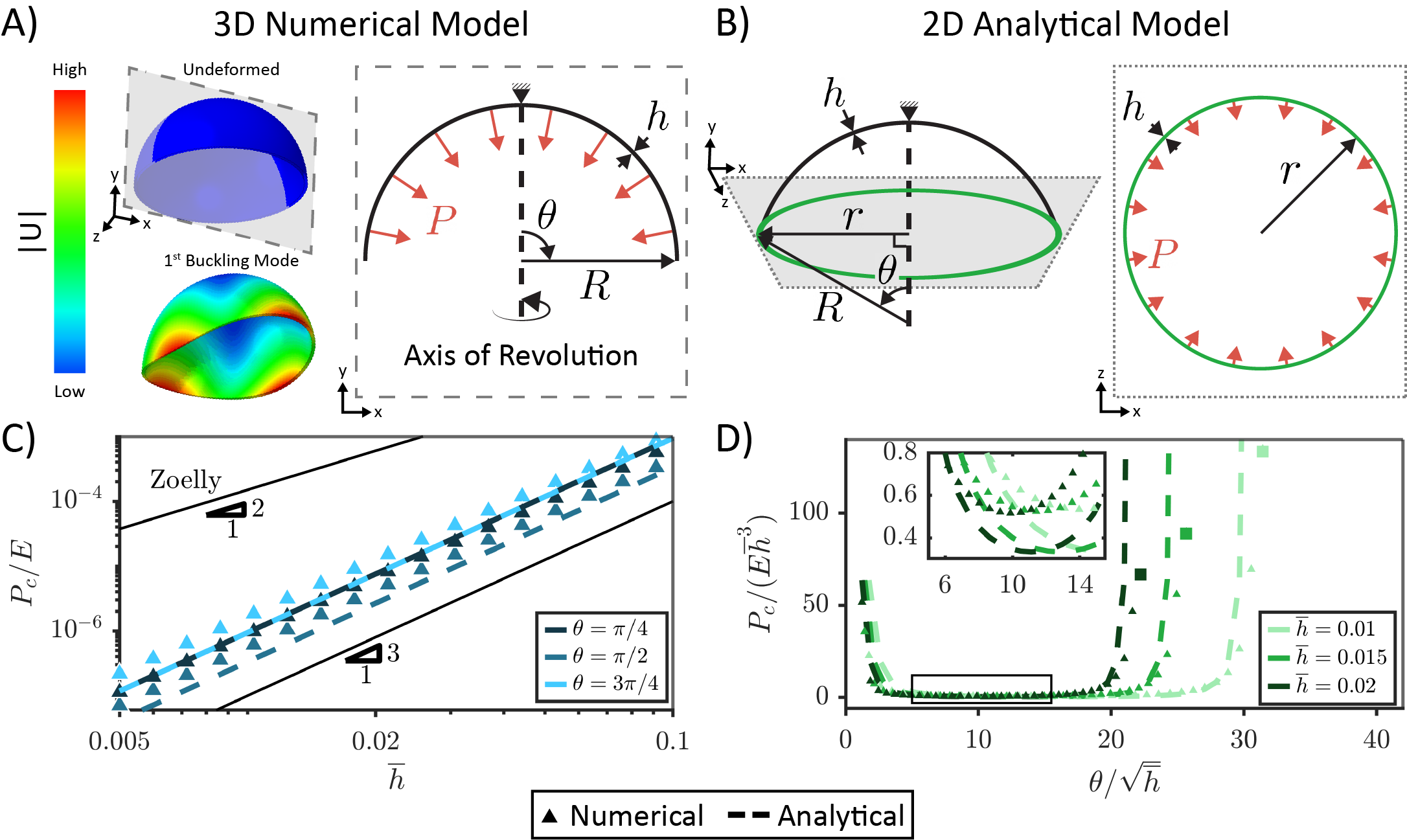}
    \caption{\textbf{Buckling Instability in Free Thin-Shell Domes.} \textbf{A)}  3D numerical model:  underformed shape and first buckling mode of the free hemispherical shell with radius $R$, thickness $h$, cap angle $\theta$, and applied internal pressure $P$. \textbf{B)}  Two-dimensional analytical model of the hemispherical shell based on the boundary edge of the spherical dome.  \textbf{C)} and \textbf{D)}  Analytical and numerical results for the normalized critical buckling pressure with respect to slenderness ratio $\bar{h
    }$ (C) and normalized cap angle $\theta/\sqrt{\bar{h}}$ (D). The squares  represent the classical buckling pressure for a full sphere under uniform pressure \cite{zoelly_ueber_1915}.}
    \label{fig:f2}
\end{figure*}
The investigation of the ``Pac-Man'' gripper first started with a finite element (FE) model to predict the buckling behavior of a free, thin hemispherical shell with radius $R$,  thickness $h$, and cap angle $\theta$, subjected to a uniform internal pressure $P$ (Fig. \ref{fig:f2}A).
In the FE analyses (performed with the commercial package ABAQUS 2020/Standard), we modeled the hemisphere using four-node shell elements (S4R) and applied a neo-Hookean material model with the mechanical properties of the rubber material Elite Double 32 from Zhermack (see Supplementary Materials ``Section 1. Finite Element Simulations" for details). 
Next, we conducted buckling analyses to extract the critical buckling pressure and corresponding mode shape for multiple values of slenderness ratio $\overline{h}=h/R$ and cap angle $\theta$ (for more information, see Supplementary Materials ``Section 1. Finite Element Simulations" for details).

Based on the FE analyses, the first buckling mode exhibits a saddle shape with two pairs of upper and lower lobes, which results in the loss of rotational symmetry of the hemispherical shell (Fig. \ref{fig:f2}A).

We noted that the in-plane deformation of the equatorial line of the hemispherical mode shape resembles the lobed, elliptical first mode of a circular ring \cite{timoshenko_theory_2009}.
Taking advantage of this observation, we developed an equivalent two-dimensional analytical model to predict the critical pressure of the rotational symmetry-breaking buckling for the hemispherical shell.
Thus, we approximate the hemispherical shell with the circular ring having the properties of the equatorial line of the shell (Fig. \ref{fig:f2}B).
The buckling pressure of a circular ring can be written as \cite{timoshenko_theory_2009}
\begin{equation}
    P_c=\frac{E}{4(1-\nu^2)}\left(\frac{h}{r}\right)^3
    \label{eq:archPress}
\end{equation}
where $E$ is the Young's modulus, $\nu$ is the Poisson's ratio, and $r$ and $h$ are the ring radius and thickness, respectively.
In our analogy, the circular ring represents the base ring of the spherical dome with cap angle $\theta$ (Fig. \ref{fig:f2}B), such that the slenderness ratio can be expressed using a projection of the hemispherical radius
\begin{equation}
    \frac{h}{r}=\frac{h}{R} \csc{\theta}
    \label{eq:radProj}
\end{equation}
By substituting Eq. \ref{eq:radProj} to Eq. \ref{eq:archPress}, we obtain the critical buckling pressure $P_c$ predicted by the 2D analytical model for the spherical dome
\begin{equation}
    P_c=\frac{E}{4(1-\nu^2)}\left(\frac{h}{R}\right)^3 \csc^3{\theta}
    \label{eq:archPressProj}
\end{equation}
We found good agreement between the critical buckling pressures predicted from the analytical model and our FE results (Fig. \ref{fig:f2}C - D).
In Fig. \ref{fig:f2}C, we demonstrate an observed cubic relationship between the critical buckling pressure and the slenderness ratio $\bar{h} = h/R$. for free hemispherical shells.
This cubic trend differs from the classical Zoelly quadratic relationship  \cite{zoelly_ueber_1915} for thin-shell hemispheres with clamped boundary conditions. The lower buckling threshold observed with respect to the Zoelly theory is the result of the near-isometric deformation that the free, thin, elastic shell can reconfigure into.
In Fig. \ref{fig:f2}D, we report the nondimensional critical pressure $ P_c / (E \bar{h}^3) $ as a function of the nondimensional cap angle $\theta / \sqrt{\bar{h}}$  and we show that both analytical and numerical results follow similar trends with a sharp increase in slope near the minimal and maximal values of the cap angle, where the geometry of the hemispherical shell approaches that of a circular plate and of a full spherical shell, respectively.
The numerical results for $\theta = \pi$ also demonstrate good agreement with the classical critical buckling pressure for full spherical shells \cite{zoelly_ueber_1915} (indicated by the square markers in Fig. \ref{fig:f2}D).

\subsection{Rational Design of the ``Pac-Man'' Gripper}
\begin{figure*}
    \centering
    \includegraphics[width=\textwidth,keepaspectratio]{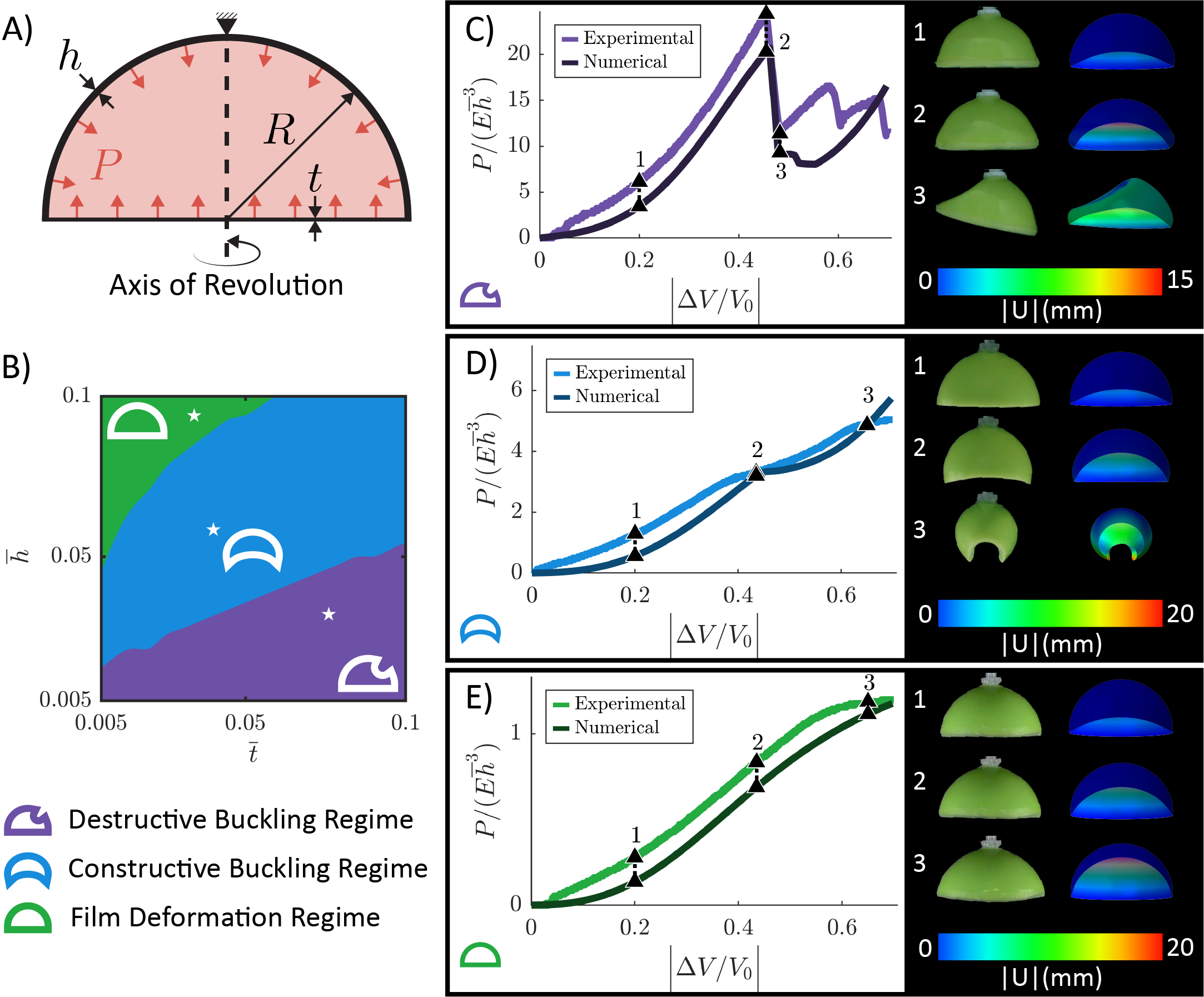}
    \caption{\textbf{Design Space and Experimental Validation.} \textbf{A)} Schematic of the hemispherical soft pneumatic actuator. \textbf{B)} Evolution of the three behavioral regimes (\textit{i.e.} `destructive' buckling, `constructive' buckling, and film deformation regime) as a function of  the slenderness ratio and normalized film thickness $\bar{h}$ and $\bar{t}$, respectively. The stars indicate the geometrical parameters of the experimental samples. \textbf{C)}, \textbf{D)} and \textbf{E)} Numerical and experimental comparison of the normalized pressure-volume curves  and deformed shape for a design in Regime 1, Regime 2, and Regime 3, respectively.}
    \label{fig:f3}
\end{figure*}

Towards this goal, we explored a fully soft gripper made of a hemispherical shell with a thin film covering the equatorial plane (Fig. \ref{fig:f3}A).
The introduction of this elastic film creates an isolated fluid cavity within the dome, which enables the use of pneumatic input to actuate the hemisphere.
We then investigated the soft gripper's response upon deflation ($P>0$ as shown in Fig. \ref{fig:f3}A) of the internal cavity both numerically and experimentally.
To study the nonlinear deformation behavior of the enclosed dome, we used the dynamic implicit solver in ABAQUS 2020/Standard for FE volume-controlled simulations by applying an incompressible fluid within the cavity (see the Supplementary Materials ``Section S1. Finite Element Simulations" for details).
The highly compliant Ecoflex 00-30 was chosen for the thin film, which we modeled using an incompressible Gent material model \cite{gent1996new} (see the Supplementary Materials ``Section S1. Finite Element Simulations" for details).
We explored the gripper's parameter space by fixing the cap angle $\theta = \pi/2$ and varying both the normalized film thickness $\overline{t}=t/R$ and the slenderness ratio over a range spanning $0.005<\overline{t}<0.1$ and $0.005<\overline{h}<0.1$.
For each set of geometric parameters, we obtained the corresponding dome's pressure-volume deflation curve and nonlinear deformed shape, which we used to classify the structures' responses (see the Supplementary Materials ``Section S1. Finite Element Simulations" for details). 

Upon deflation, we found that the behavior of the soft actuator splits into three distinct regimes (Fig. \ref{fig:f3}B): (\textit{i}) the `destructive' buckling regime (Regime 1), (\textit{ii}) the `constructive' buckling regime (Regime 2), and (\textit{iii}) the film deformation regime (Regime 3).
The `destructive' buckling regime occurs for geometries with high film thickness and low shell thickness.
In these geometries, the film stiffness outweighs the shell stiffness, such that the film acts as a fixed boundary on the shell.
This boundary effect, highlighted in Fig. \ref{fig:f3}C, gives rise to a shell buckling response typical of soft spherical shells with clamped boundary conditions \cite{lee2016geometric, hutchinson2017nonlinear}.
In Fig. \ref{fig:f3}C, the numerical pressure-volume curve of the hemisphere with slenderness ratio $\overline{h}=0.032$ and normalized film thickness $\overline{t}=0.076$  shows the classic sudden drop in pressure that corresponds to the catastrophic buckling onset for $\left|\Delta V/V_0=0.455\right|$.
This traditional, catastrophic shell buckling response does not produce a mode shape conducive for grasping or encapsulation, so we categorize the shells in this regime as undergoing `destructive', or nonfunctional, buckling.

The `constructive' buckling regime (Fig. \ref{fig:f3}D) arises when the film and shell thicknesses are comparable and both structural elements deform in unison.
Here, the dome's reconfiguration closely aligns with the mode shape of the hemispherical shell with no applied film (Fig. \ref{fig:f2}A). In Fig. \ref{fig:f3}D, the predicted FE pressure-volume curve of the hemisphere with slenderness ratio $\overline{h}=0.058$ and normalized film thickness $\overline{t}=0.040$ shows a sudden change in slope at the onset of the rotational symmetry-breaking buckling for $\left|\Delta V/V_0 =0.435\right|$.
By triggering this instability, which pinches together the two lobes on the minor axis, the shell's deformation facilitates secure grasping and encapsulation.
The shells in this regime that exhibit the ``Pac-Man'' mode shape and they are the most functional for manipulation tasks, so we classify them as undergoing `constructive' buckling.

The third and final regime, referred to as the film deformation regime, is governed by excessive deformation of the film surface and minimal deformation of the shell.
It is found for high values of shell thickness and low values of film thickness (Fig. \ref{fig:f3}E).
In Fig. \ref{fig:f3}E, the numerical pressure-volume curve of the hemisphere with slenderness ratio $\overline{h}=0.094$ and normalized film thickness  $\overline{t}=0.034$  presents a fairly linear trend without any appearance of a buckling instability.
Similarly to the `destructive' buckling regime, domes that fall within the film deformation regime do not produce significant enough deformation of the equatorial line to be harnessed to grasping tasks.
These observations about the viability of each regime for grasping effectively reduces the design space of feasible ``Pac-Man" grippers to the `constructive' buckling region bounded in blue within Fig. \ref{fig:f3}B.

Inspired by our numerical findings, we experimentally validated our FE analyses by selecting three geometric configurations, one from each regime.
The elastic samples were manufactured with a combination of coating \cite{lee2016fabrication} and mold-and-cast processes (see Supplementary Materials ``Section S2. Manufacturing" for more details) using Zhermack Elite Double 32 and Smooth-On Ecoflex 00-30 for the hemispherical shell and the thin film, respectively.
Experimental pressure-volume curves were obtained by deflating the hemispherical internal cavity using air as the working fluid and accounting for compressibility effects (see Supplementary Materials ``Section S3. Experimental Set-up and Testing" for more details).
In all the three regimes, the experimental responses show very good agreement with the FE results, and pressure measurements accurately match all the three numerically predicted trends (Fig. \ref{fig:f3}C - E and movie S2).

\maketitle

\subsection{Grasping with a Sense of Touch in Complex Environments}

\begin{figure*}
    \centering
    \includegraphics[width=\textwidth,keepaspectratio]{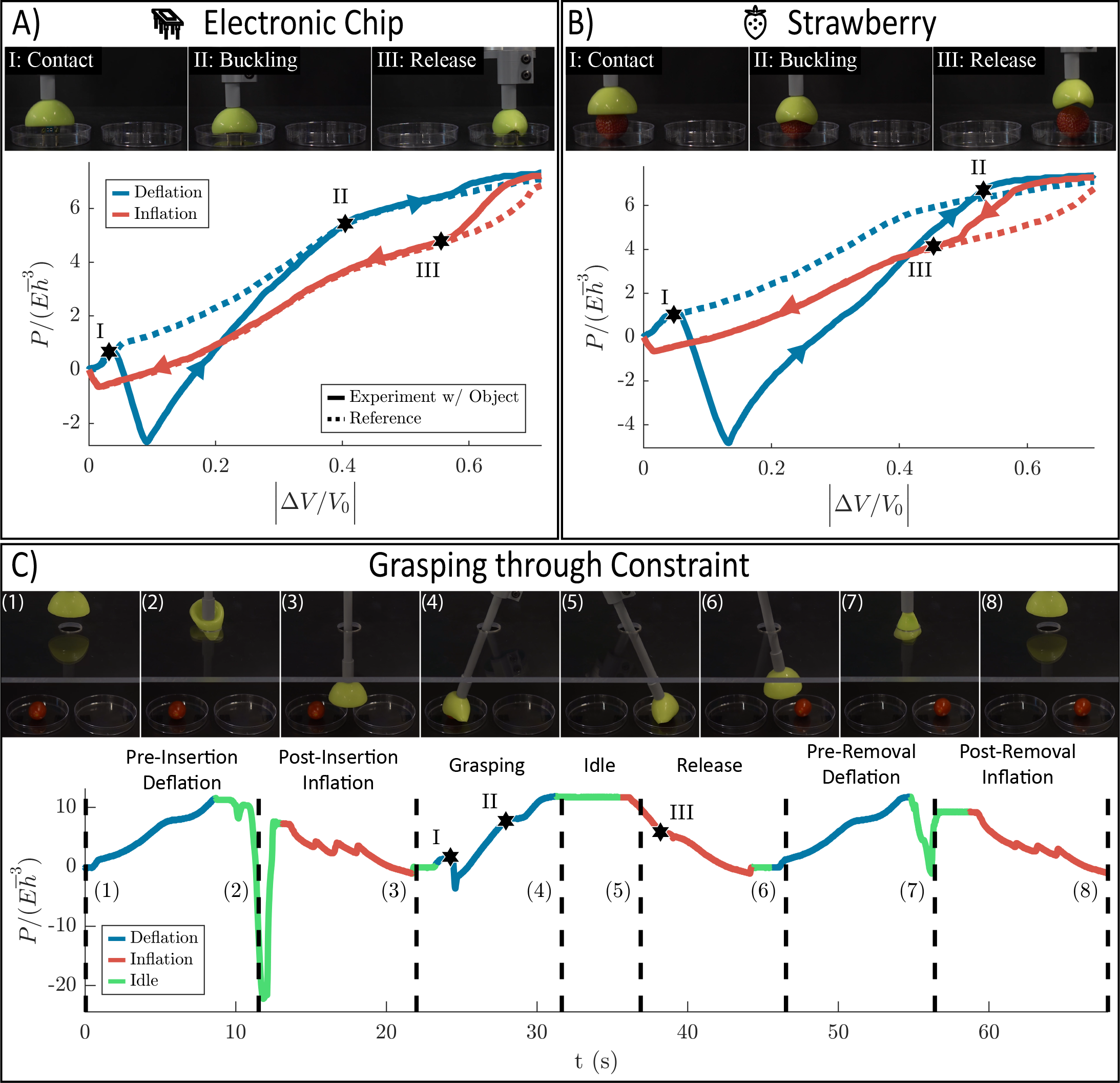}
    \caption{\textbf{Experimental Grasping Trials.} \textbf{A)} and \textbf{B)} Experimental snapshots and normalized pressure-volume curves of the ``Pac-Man'' gripper grasping an electronic chip and a strawberry, respectively. The dashed lines represent the response of the gripper without grasping an object.  \textbf{C)} Experimental snapshots and normalized pressure-time response of the ``Pac-Man'' gripper operating through a constrained opening to manipulate a cherry tomato. (I) Contact, (II) Buckling, and (III) Release key points are labeled in all the experimental curves.
    }
    \label{fig:f4}
\end{figure*}

Having demonstrated the boundaries of the constructive buckling regime, ideal for grasping tasks, we then tested the ``Pac-Man" gripper's ability to manipulate objects of varying shapes, textures, slipperiness, and stiffness.
In this experimental analysis, we connected the ``Pac-Man" gripper and the pneumatic system to a commercial robotic arm to precisely and reliably control the motion of the soft actuator.

Fig. \ref{fig:f4}A - B show the soft gripper ($\bar{h} = 0.052 $ and $\bar{t} = 0.048 $) grasping and releasing an electronic chip and a strawberry, respectively, which vary strongly in size, shape, and fragility.
The versatility of the ``Pac-Man" gripper is also demonstrated by the grasping of multiple different objects in movie S3.
First, we collected baseline pressure-volume curves from the gripper without grasping any object.
These reference curves were then overlapped with the experimental curves of the gripper recorded while performing a grasping task.
By highlighting differences between the reference and experimental responses, we can locate meaningful points that characterize the interaction of the gripper with the grasped object.
Thus, we developed an algorithm capable of identifying the moment of contact, buckling (i.e. grasping), and release of the objects in real time (see Supplementary Materials ``Section S3. Experimental Set-up and Testing" for details).
From each deflation curve, a significant downward jump can be seen as the gripper contacts the object and, consequently, the absolute pressure in the fluid cavity increases.
In Fig. \ref{fig:f4}B, the strawberry imposes a larger deformation on the film with respect to the electronic chip, which results in a larger negative pressure spike after contact.
When a large change in slope of the deflation pressure-volume curve is detected, we observe buckling, which signals the onset of the secure grasping ``Pac-Man" deformation.
During re-inflation, the experimental and reference curves overlap at the point when the object is released from grip, enabling us to identify the instant of loss of contact.
These experiments demonstrate the power of the buckling-based hemispherical gripper by highlighting the simple, integrated sensing capabilities observed from the pressure-volume response alone.

The combination of these advanced functionalities with the gripper's extreme compliance also enables a wider range of operations such as grasping objects in confined environments.
Fig. \ref{fig:f4}C depicts the gripper ($\bar{h} = 0.044 $ and $\bar{t} = 0.048 $) being used to perform a grasping task through a constrained opening in a rigid plate.
The gripper is deflated to fit within the confined space, and is then reinflated to perform the grasping task.
Once the task is completed, the gripper is deflated once again to be removed from the confined space, where it can be reinflated back to its initial configuration.
Through the pressure measurements recorded during the deflation, idle, and inflation phases, we can identify the interactions of the ``Pac-Man" gripper not only with the grasped objects but also with the surrounding environment.
The pressure curve gives the opportunity to classify of the contact, grasping, and release points with the object, as well as to identify its passage through the confined opening.
Indeed, during the idle phase, a large spike is observed when the soft gripper is inserted into ($t=10.8$ s) and removed from ($t=54.8$ s) the tiny hole, respectively.
During insertion, the pressure spike is deeper because the motion is opposite to the natural curvature of the soft gripping shell.
However, the spike is smaller when the soft gripper is being removed, since it can fold more easily into a smaller diameter shape.
Hence, the amplitude of the pressure spike can be used to infer the complexity of the interaction with the surrounding environment.
The response during this constrained experiment and during an additional multi-step grasping trial can be found in movie S4.

\section{Conclusion}
In summary, we have introduced a novel class of hemispherical gripper, named ``Pac-Man'' gripper, capable of manipulating fragile, slippery, and deformable bodies, while simultaneously providing a simple, embedded, responsive electronics-free sensing strategy.
Enabled by a buckling instability, this gripper reconfigures rapidly into a two-node mode shape ideal for grasping and encapsulating objects even within narrow and confined environments.
We investigated the design principles that can guide the development of grippers that demonstrate this buckling phenomenon, which we harnessed to manipulate objects ranging in fragility, size, and shape.
The resulting large amplitude reconfigurations of the device enable it to perform complex grasping tasks after being forced through openings smaller than the size of the gripper itself.
Since the buckling instability is scale independent, the ``Pac-Man'' gripper's grasping behavior, here demonstrated at the centimeter scale, creates opportunities to develop large or micro scale grippers based on the same fundamental principles.

The inherent simplicity of the design enables to collect nuanced sensing information  purely by monitoring the pressure-volume response of the soft gripper while interacting with the grasped objects and the confined surrounding environment. Contact, buckling, and release key points appear as discernible inflection points in the pressure-volume collected data and they can be located real-time.
This data processing is performed in real time, creating countless opportunities to utilize the device in constrained, extreme, and low visibility environments without the need for additional sensing components.
These features combine to yield a simple, comprehensive solution to universal grasping that includes the embedded sensing capability to extract meaningful information about live interactions across any physical scale.

Variations of the proposed design can integrate remote control using stimuli-responsive materials \cite{lee2021stimuli}, such as liquid crystal elastomer shells \cite{jampani2019liquid} and magneto-active elastic shells \cite{yan2021magneto}. Localized shell imperfections \cite{lee2016geometric} can also be seeded and harnessed to tune the ``Pac-Man'' shell buckling onset and preprogram the gripper response. We anticipate that this highly-responsive and scalable soft gripper  will set a new path for the next-generation haptic devices and soft robots with electronics-free sense of touch designed to operate in unpredictable environments.

\section{Materials and Methods}
Rigorous descriptions of the methodology used for each step of our investigation are provided within the Supplementary Materials.
The design of the hemispherical grippers as well as the manufacturing process we utilized are described in Section S1.
The experimental setup we utilized, the steps we took to account for compressibility, a discussion of the gripper's payload-to-weight ratio, and the methodology we adopted to label key points from pressure data are detailed in Section S2.
Parameters used for the Finite Element Method investigations and the process we leveraged to categorize shells within our design space are explored in Section S3.

\section{Acknowledgements}
We gratefully acknowledge the support of the Maryland Robotics Center (MRC) as well as the Robotics Manipulator Lab at UMD.

\textbf{Funding:} E.T. was supported by funding from the University of Maryland Start-up Package. K.B. was supported by funding from the Maryland Robotics Center Pathway to PhD program as well as the Clark Doctoral Fellowship from the University of Maryland.

\textbf{Author Contributions:} E.T., M.P., and G.L. proposed the research concept. K.B. conducted modeling and numerical simulations. Z.C. manufactured all experimental samples. K.B. and Z.C. conducted experimental work. K.B. and E.T. analyzed the data and wrote the paper. All co-authors revised the paper. E.T. supervised and coordinated the research.

\textbf{Competing Interests:} The authors declare that they do not have any competing interests.

\textbf{Data and Materials Availability:} All data necessary to evaluate the conclusions of the paper are shown within the main text or the Supplementary Materials.

\bibliography{Bibliography/biblios}

\begin{thebibliography}{59}%
\makeatletter
\providecommand \@ifxundefined [1]{%
 \@ifx{#1\undefined}
}%
\providecommand \@ifnum [1]{%
 \ifnum #1\expandafter \@firstoftwo
 \else \expandafter \@secondoftwo
 \fi
}%
\providecommand \@ifx [1]{%
 \ifx #1\expandafter \@firstoftwo
 \else \expandafter \@secondoftwo
 \fi
}%
\providecommand \natexlab [1]{#1}%
\providecommand \enquote  [1]{``#1''}%
\providecommand \bibnamefont  [1]{#1}%
\providecommand \bibfnamefont [1]{#1}%
\providecommand \citenamefont [1]{#1}%
\providecommand \href@noop [0]{\@secondoftwo}%
\providecommand \href [0]{\begingroup \@sanitize@url \@href}%
\providecommand \@href[1]{\@@startlink{#1}\@@href}%
\providecommand \@@href[1]{\endgroup#1\@@endlink}%
\providecommand \@sanitize@url [0]{\catcode `\\12\catcode `\$12\catcode `\&12\catcode `\#12\catcode `\^12\catcode `\_12\catcode `\%12\relax}%
\providecommand \@@startlink[1]{}%
\providecommand \@@endlink[0]{}%
\providecommand \url  [0]{\begingroup\@sanitize@url \@url }%
\providecommand \@url [1]{\endgroup\@href {#1}{\urlprefix }}%
\providecommand \urlprefix  [0]{URL }%
\providecommand \Eprint [0]{\href }%
\providecommand \doibase [0]{https://doi.org/}%
\providecommand \selectlanguage [0]{\@gobble}%
\providecommand \bibinfo  [0]{\@secondoftwo}%
\providecommand \bibfield  [0]{\@secondoftwo}%
\providecommand \translation [1]{[#1]}%
\providecommand \BibitemOpen [0]{}%
\providecommand \bibitemStop [0]{}%
\providecommand \bibitemNoStop [0]{.\EOS\space}%
\providecommand \EOS [0]{\spacefactor3000\relax}%
\providecommand \BibitemShut  [1]{\csname bibitem#1\endcsname}%
\let\auto@bib@innerbib\@empty
\bibitem [{\citenamefont {Thompson}(1992)}]{thompson_growth_1992}%
  \BibitemOpen
  \bibfield  {author} {\bibinfo {author} {\bibfnamefont {D.~W.}\ \bibnamefont {Thompson}},\ }\href {https://doi.org/10.1017/CBO9781107325852} {\emph {\bibinfo {title} {On Growth and Form}}},\ \bibinfo {edition} {1st}\ ed.,\ edited by\ \bibinfo {editor} {\bibfnamefont {J.~T.}\ \bibnamefont {Bonner}}\ (\bibinfo  {publisher} {Cambridge University Press},\ \bibinfo {year} {1992})\BibitemShut {NoStop}%
\bibitem [{\citenamefont {Ball}(2009)}]{ball_natures_2009}%
  \BibitemOpen
  \bibfield  {author} {\bibinfo {author} {\bibfnamefont {P.}~\bibnamefont {Ball}},\ }\href@noop {} {\emph {\bibinfo {title} {Nature's patterns: a tapestry in three parts}}}\ (\bibinfo  {publisher} {Oxford University Press},\ \bibinfo {year} {2009})\ \bibinfo {note} {{OCLC}: ocn729742295}\BibitemShut {NoStop}%
\bibitem [{\citenamefont {Banavar}\ \emph {et~al.}(2014)\citenamefont {Banavar}, \citenamefont {Cooke}, \citenamefont {Rinaldo},\ and\ \citenamefont {Maritan}}]{banavar_form_2014}%
  \BibitemOpen
  \bibfield  {author} {\bibinfo {author} {\bibfnamefont {J.~R.}\ \bibnamefont {Banavar}}, \bibinfo {author} {\bibfnamefont {T.~J.}\ \bibnamefont {Cooke}}, \bibinfo {author} {\bibfnamefont {A.}~\bibnamefont {Rinaldo}},\ and\ \bibinfo {author} {\bibfnamefont {A.}~\bibnamefont {Maritan}},\ }\bibfield  {title} {\bibinfo {title} {Form, function, and evolution of living organisms},\ }\href@noop {} {\bibfield  {journal} {\bibinfo  {journal} {Proceedings of the National Academy of Sciences}\ }\textbf {\bibinfo {volume} {111}},\ \bibinfo {pages} {3332} (\bibinfo {year} {2014})}\BibitemShut {NoStop}%
\bibitem [{\citenamefont {Yang}\ \emph {et~al.}(2021)\citenamefont {Yang}, \citenamefont {Vella},\ and\ \citenamefont {Holmes}}]{yang_grasping_2021}%
  \BibitemOpen
  \bibfield  {author} {\bibinfo {author} {\bibfnamefont {Y.}~\bibnamefont {Yang}}, \bibinfo {author} {\bibfnamefont {K.}~\bibnamefont {Vella}},\ and\ \bibinfo {author} {\bibfnamefont {D.~P.}\ \bibnamefont {Holmes}},\ }\bibfield  {title} {\bibinfo {title} {Grasping with kirigami shells},\ }\href@noop {} {\bibfield  {journal} {\bibinfo  {journal} {Science Robotics}\ }\textbf {\bibinfo {volume} {6}},\ \bibinfo {pages} {eabd6426} (\bibinfo {year} {2021})}\BibitemShut {NoStop}%
\bibitem [{\citenamefont {Hong}\ \emph {et~al.}(2022)\citenamefont {Hong}, \citenamefont {Chi}, \citenamefont {Wu}, \citenamefont {Li}, \citenamefont {Zhu},\ and\ \citenamefont {Yin}}]{hong_boundary_2022}%
  \BibitemOpen
  \bibfield  {author} {\bibinfo {author} {\bibfnamefont {Y.}~\bibnamefont {Hong}}, \bibinfo {author} {\bibfnamefont {Y.}~\bibnamefont {Chi}}, \bibinfo {author} {\bibfnamefont {S.}~\bibnamefont {Wu}}, \bibinfo {author} {\bibfnamefont {Y.}~\bibnamefont {Li}}, \bibinfo {author} {\bibfnamefont {Y.}~\bibnamefont {Zhu}},\ and\ \bibinfo {author} {\bibfnamefont {J.}~\bibnamefont {Yin}},\ }\bibfield  {title} {\bibinfo {title} {Boundary curvature guided programmable shape-morphing kirigami sheets},\ }\href@noop {} {\bibfield  {journal} {\bibinfo  {journal} {Nature Communications}\ }\textbf {\bibinfo {volume} {13}},\ \bibinfo {pages} {530} (\bibinfo {year} {2022})}\BibitemShut {NoStop}%
\bibitem [{\citenamefont {Brown}\ \emph {et~al.}(2010)\citenamefont {Brown}, \citenamefont {Rodenberg}, \citenamefont {Amend}, \citenamefont {Mozeika}, \citenamefont {Steltz}, \citenamefont {Zakin}, \citenamefont {Lipson},\ and\ \citenamefont {Jaeger}}]{brown_universal_2010}%
  \BibitemOpen
  \bibfield  {author} {\bibinfo {author} {\bibfnamefont {E.}~\bibnamefont {Brown}}, \bibinfo {author} {\bibfnamefont {N.}~\bibnamefont {Rodenberg}}, \bibinfo {author} {\bibfnamefont {J.}~\bibnamefont {Amend}}, \bibinfo {author} {\bibfnamefont {A.}~\bibnamefont {Mozeika}}, \bibinfo {author} {\bibfnamefont {E.}~\bibnamefont {Steltz}}, \bibinfo {author} {\bibfnamefont {M.~R.}\ \bibnamefont {Zakin}}, \bibinfo {author} {\bibfnamefont {H.}~\bibnamefont {Lipson}},\ and\ \bibinfo {author} {\bibfnamefont {H.~M.}\ \bibnamefont {Jaeger}},\ }\bibfield  {title} {\bibinfo {title} {Universal robotic gripper based on the jamming of granular material},\ }\href@noop {} {\bibfield  {journal} {\bibinfo  {journal} {Proceedings of the National Academy of Sciences}\ }\textbf {\bibinfo {volume} {107}},\ \bibinfo {pages} {18809} (\bibinfo {year} {2010})}\BibitemShut {NoStop}%
\bibitem [{\citenamefont {Li}\ \emph {et~al.}(2019)\citenamefont {Li}, \citenamefont {Stampfli}, \citenamefont {Xu}, \citenamefont {Malkin}, \citenamefont {Diaz}, \citenamefont {Rus},\ and\ \citenamefont {Wood}}]{li_vacuum-driven_2019}%
  \BibitemOpen
  \bibfield  {author} {\bibinfo {author} {\bibfnamefont {S.}~\bibnamefont {Li}}, \bibinfo {author} {\bibfnamefont {J.~J.}\ \bibnamefont {Stampfli}}, \bibinfo {author} {\bibfnamefont {H.~J.}\ \bibnamefont {Xu}}, \bibinfo {author} {\bibfnamefont {E.}~\bibnamefont {Malkin}}, \bibinfo {author} {\bibfnamefont {E.~V.}\ \bibnamefont {Diaz}}, \bibinfo {author} {\bibfnamefont {D.}~\bibnamefont {Rus}},\ and\ \bibinfo {author} {\bibfnamefont {R.~J.}\ \bibnamefont {Wood}},\ }\bibfield  {title} {\bibinfo {title} {A vacuum-driven origami “magic-ball” soft gripper},\ }in\ \href@noop {} {\emph {\bibinfo {booktitle} {International Conference on Robotics and Automation}}}\ (\bibinfo {organization} {IEEE},\ \bibinfo {year} {2019})\ pp.\ \bibinfo {pages} {7401--7408}\BibitemShut {NoStop}%
\bibitem [{\citenamefont {Becker}\ \emph {et~al.}(2022)\citenamefont {Becker}, \citenamefont {Teeple}, \citenamefont {Charles}, \citenamefont {Jung}, \citenamefont {Baum}, \citenamefont {Weaver}, \citenamefont {Mahadevan},\ and\ \citenamefont {Wood}}]{becker_active_2022}%
  \BibitemOpen
  \bibfield  {author} {\bibinfo {author} {\bibfnamefont {K.}~\bibnamefont {Becker}}, \bibinfo {author} {\bibfnamefont {C.}~\bibnamefont {Teeple}}, \bibinfo {author} {\bibfnamefont {N.}~\bibnamefont {Charles}}, \bibinfo {author} {\bibfnamefont {Y.}~\bibnamefont {Jung}}, \bibinfo {author} {\bibfnamefont {D.}~\bibnamefont {Baum}}, \bibinfo {author} {\bibfnamefont {J.~C.}\ \bibnamefont {Weaver}}, \bibinfo {author} {\bibfnamefont {L.}~\bibnamefont {Mahadevan}},\ and\ \bibinfo {author} {\bibfnamefont {R.}~\bibnamefont {Wood}},\ }\bibfield  {title} {\bibinfo {title} {Active entanglement enables stochastic, topological grasping},\ }\href@noop {} {\bibfield  {journal} {\bibinfo  {journal} {Proceedings of the National Academy of Sciences}\ }\textbf {\bibinfo {volume} {119}},\ \bibinfo {pages} {e2209819119} (\bibinfo {year} {2022})}\BibitemShut {NoStop}%
\bibitem [{\citenamefont {Sinatra}\ \emph {et~al.}(2019)\citenamefont {Sinatra}, \citenamefont {Teeple}, \citenamefont {Vogt}, \citenamefont {Parker}, \citenamefont {Gruber},\ and\ \citenamefont {Wood}}]{sinatra_ultragentle_2019}%
  \BibitemOpen
  \bibfield  {author} {\bibinfo {author} {\bibfnamefont {N.~R.}\ \bibnamefont {Sinatra}}, \bibinfo {author} {\bibfnamefont {C.~B.}\ \bibnamefont {Teeple}}, \bibinfo {author} {\bibfnamefont {D.~M.}\ \bibnamefont {Vogt}}, \bibinfo {author} {\bibfnamefont {K.~K.}\ \bibnamefont {Parker}}, \bibinfo {author} {\bibfnamefont {D.~F.}\ \bibnamefont {Gruber}},\ and\ \bibinfo {author} {\bibfnamefont {R.~J.}\ \bibnamefont {Wood}},\ }\bibfield  {title} {\bibinfo {title} {Ultragentle manipulation of delicate structures using a soft robotic gripper},\ }\href@noop {} {\bibfield  {journal} {\bibinfo  {journal} {Science Robotics}\ }\textbf {\bibinfo {volume} {4}},\ \bibinfo {pages} {eaax5425} (\bibinfo {year} {2019})}\BibitemShut {NoStop}%
\bibitem [{\citenamefont {Yoder}\ \emph {et~al.}(2023)\citenamefont {Yoder}, \citenamefont {Macari}, \citenamefont {Kleinwaks}, \citenamefont {Schmidt}, \citenamefont {Acome},\ and\ \citenamefont {Keplinger}}]{yoder_soft_2023}%
  \BibitemOpen
  \bibfield  {author} {\bibinfo {author} {\bibfnamefont {Z.}~\bibnamefont {Yoder}}, \bibinfo {author} {\bibfnamefont {D.}~\bibnamefont {Macari}}, \bibinfo {author} {\bibfnamefont {G.}~\bibnamefont {Kleinwaks}}, \bibinfo {author} {\bibfnamefont {I.}~\bibnamefont {Schmidt}}, \bibinfo {author} {\bibfnamefont {E.}~\bibnamefont {Acome}},\ and\ \bibinfo {author} {\bibfnamefont {C.}~\bibnamefont {Keplinger}},\ }\bibfield  {title} {\bibinfo {title} {A soft, fast and versatile electrohydraulic gripper with capacitive object size detection},\ }\href@noop {} {\bibfield  {journal} {\bibinfo  {journal} {Advanced Functional Materials}\ }\textbf {\bibinfo {volume} {33}},\ \bibinfo {pages} {2209080} (\bibinfo {year} {2023})}\BibitemShut {NoStop}%
\bibitem [{\citenamefont {Truby}\ \emph {et~al.}(2018)\citenamefont {Truby}, \citenamefont {Wehner}, \citenamefont {Grosskopf}, \citenamefont {Vogt}, \citenamefont {Uzel}, \citenamefont {Wood},\ and\ \citenamefont {Lewis}}]{truby_soft_2018}%
  \BibitemOpen
  \bibfield  {author} {\bibinfo {author} {\bibfnamefont {R.~L.}\ \bibnamefont {Truby}}, \bibinfo {author} {\bibfnamefont {M.}~\bibnamefont {Wehner}}, \bibinfo {author} {\bibfnamefont {A.~K.}\ \bibnamefont {Grosskopf}}, \bibinfo {author} {\bibfnamefont {D.~M.}\ \bibnamefont {Vogt}}, \bibinfo {author} {\bibfnamefont {S.~G.}\ \bibnamefont {Uzel}}, \bibinfo {author} {\bibfnamefont {R.~J.}\ \bibnamefont {Wood}},\ and\ \bibinfo {author} {\bibfnamefont {J.~A.}\ \bibnamefont {Lewis}},\ }\bibfield  {title} {\bibinfo {title} {Soft somatosensitive actuators via embedded 3d printing},\ }\href@noop {} {\bibfield  {journal} {\bibinfo  {journal} {Advanced Materials}\ }\textbf {\bibinfo {volume} {30}},\ \bibinfo {pages} {1706383} (\bibinfo {year} {2018})}\BibitemShut {NoStop}%
\bibitem [{\citenamefont {Li}\ \emph {et~al.}(2022)\citenamefont {Li}, \citenamefont {Huh}, \citenamefont {Yahnker},\ and\ \citenamefont {Stuart}}]{li_resonant_2022}%
  \BibitemOpen
  \bibfield  {author} {\bibinfo {author} {\bibfnamefont {M.~S.}\ \bibnamefont {Li}}, \bibinfo {author} {\bibfnamefont {T.~M.}\ \bibnamefont {Huh}}, \bibinfo {author} {\bibfnamefont {C.~R.}\ \bibnamefont {Yahnker}},\ and\ \bibinfo {author} {\bibfnamefont {H.~S.}\ \bibnamefont {Stuart}},\ }\bibfield  {title} {\bibinfo {title} {Resonant pneumatic tactile sensing for soft grippers},\ }\href@noop {} {\bibfield  {journal} {\bibinfo  {journal} {IEEE Robotics and Automation Letters}\ }\textbf {\bibinfo {volume} {7}},\ \bibinfo {pages} {10105} (\bibinfo {year} {2022})}\BibitemShut {NoStop}%
\bibitem [{\citenamefont {Acome}\ \emph {et~al.}(2018)\citenamefont {Acome}, \citenamefont {Mitchell}, \citenamefont {Morrissey}, \citenamefont {Emmett}, \citenamefont {Benjamin}, \citenamefont {King}, \citenamefont {Radakovitz},\ and\ \citenamefont {Keplinger}}]{acome_hydraulically_2018}%
  \BibitemOpen
  \bibfield  {author} {\bibinfo {author} {\bibfnamefont {E.}~\bibnamefont {Acome}}, \bibinfo {author} {\bibfnamefont {S.~K.}\ \bibnamefont {Mitchell}}, \bibinfo {author} {\bibfnamefont {T.}~\bibnamefont {Morrissey}}, \bibinfo {author} {\bibfnamefont {M.}~\bibnamefont {Emmett}}, \bibinfo {author} {\bibfnamefont {C.}~\bibnamefont {Benjamin}}, \bibinfo {author} {\bibfnamefont {M.}~\bibnamefont {King}}, \bibinfo {author} {\bibfnamefont {M.}~\bibnamefont {Radakovitz}},\ and\ \bibinfo {author} {\bibfnamefont {C.}~\bibnamefont {Keplinger}},\ }\bibfield  {title} {\bibinfo {title} {Hydraulically amplified self-healing electrostatic actuators with muscle-like performance},\ }\href@noop {} {\bibfield  {journal} {\bibinfo  {journal} {Science}\ }\textbf {\bibinfo {volume} {359}},\ \bibinfo {pages} {61} (\bibinfo {year} {2018})}\BibitemShut {NoStop}%
\bibitem [{\citenamefont {Liu}\ \emph {et~al.}(2020)\citenamefont {Liu}, \citenamefont {Wang}, \citenamefont {Ren}, \citenamefont {Jin}, \citenamefont {Zhang}, \citenamefont {Chen},\ and\ \citenamefont {Yan}}]{liu_polyionic_2020}%
  \BibitemOpen
  \bibfield  {author} {\bibinfo {author} {\bibfnamefont {Z.}~\bibnamefont {Liu}}, \bibinfo {author} {\bibfnamefont {Y.}~\bibnamefont {Wang}}, \bibinfo {author} {\bibfnamefont {Y.}~\bibnamefont {Ren}}, \bibinfo {author} {\bibfnamefont {G.}~\bibnamefont {Jin}}, \bibinfo {author} {\bibfnamefont {C.}~\bibnamefont {Zhang}}, \bibinfo {author} {\bibfnamefont {W.}~\bibnamefont {Chen}},\ and\ \bibinfo {author} {\bibfnamefont {F.}~\bibnamefont {Yan}},\ }\bibfield  {title} {\bibinfo {title} {Poly (ionic liquid) hydrogel-based anti-freezing ionic skin for a soft robotic gripper},\ }\href@noop {} {\bibfield  {journal} {\bibinfo  {journal} {Materials Horizons}\ }\textbf {\bibinfo {volume} {7}},\ \bibinfo {pages} {919} (\bibinfo {year} {2020})}\BibitemShut {NoStop}%
\bibitem [{\citenamefont {Terryn}\ \emph {et~al.}(2017)\citenamefont {Terryn}, \citenamefont {Brancart}, \citenamefont {Lefeber}, \citenamefont {Van~Assche},\ and\ \citenamefont {Vanderborght}}]{terryn_self-healing_2017}%
  \BibitemOpen
  \bibfield  {author} {\bibinfo {author} {\bibfnamefont {S.}~\bibnamefont {Terryn}}, \bibinfo {author} {\bibfnamefont {J.}~\bibnamefont {Brancart}}, \bibinfo {author} {\bibfnamefont {D.}~\bibnamefont {Lefeber}}, \bibinfo {author} {\bibfnamefont {G.}~\bibnamefont {Van~Assche}},\ and\ \bibinfo {author} {\bibfnamefont {B.}~\bibnamefont {Vanderborght}},\ }\bibfield  {title} {\bibinfo {title} {Self-healing soft pneumatic robots},\ }\href@noop {} {\bibfield  {journal} {\bibinfo  {journal} {Science Robotics}\ }\textbf {\bibinfo {volume} {2}},\ \bibinfo {pages} {eaan4268} (\bibinfo {year} {2017})}\BibitemShut {NoStop}%
\bibitem [{\citenamefont {Yasuda}\ \emph {et~al.}(2021)\citenamefont {Yasuda}, \citenamefont {Buskohl}, \citenamefont {Gillman}, \citenamefont {Murphey}, \citenamefont {Stepney}, \citenamefont {Vaia},\ and\ \citenamefont {Raney}}]{yasuda2021mechanical}%
  \BibitemOpen
  \bibfield  {author} {\bibinfo {author} {\bibfnamefont {H.}~\bibnamefont {Yasuda}}, \bibinfo {author} {\bibfnamefont {P.~R.}\ \bibnamefont {Buskohl}}, \bibinfo {author} {\bibfnamefont {A.}~\bibnamefont {Gillman}}, \bibinfo {author} {\bibfnamefont {T.~D.}\ \bibnamefont {Murphey}}, \bibinfo {author} {\bibfnamefont {S.}~\bibnamefont {Stepney}}, \bibinfo {author} {\bibfnamefont {R.~A.}\ \bibnamefont {Vaia}},\ and\ \bibinfo {author} {\bibfnamefont {J.~R.}\ \bibnamefont {Raney}},\ }\bibfield  {title} {\bibinfo {title} {Mechanical computing},\ }\href@noop {} {\bibfield  {journal} {\bibinfo  {journal} {Nature}\ }\textbf {\bibinfo {volume} {598}},\ \bibinfo {pages} {39} (\bibinfo {year} {2021})}\BibitemShut {NoStop}%
\bibitem [{\citenamefont {Lin}\ \emph {et~al.}(2023)\citenamefont {Lin}, \citenamefont {Wang}, \citenamefont {Zhao}, \citenamefont {Xu}, \citenamefont {Wang}, \citenamefont {Zhang}, \citenamefont {Sun}, \citenamefont {Lin},\ and\ \citenamefont {Peng}}]{lin2023recent}%
  \BibitemOpen
  \bibfield  {author} {\bibinfo {author} {\bibfnamefont {Z.}~\bibnamefont {Lin}}, \bibinfo {author} {\bibfnamefont {Z.}~\bibnamefont {Wang}}, \bibinfo {author} {\bibfnamefont {W.}~\bibnamefont {Zhao}}, \bibinfo {author} {\bibfnamefont {Y.}~\bibnamefont {Xu}}, \bibinfo {author} {\bibfnamefont {X.}~\bibnamefont {Wang}}, \bibinfo {author} {\bibfnamefont {T.}~\bibnamefont {Zhang}}, \bibinfo {author} {\bibfnamefont {Z.}~\bibnamefont {Sun}}, \bibinfo {author} {\bibfnamefont {L.}~\bibnamefont {Lin}},\ and\ \bibinfo {author} {\bibfnamefont {Z.}~\bibnamefont {Peng}},\ }\bibfield  {title} {\bibinfo {title} {Recent advances in perceptive intelligence for soft robotics},\ }\href@noop {} {\bibfield  {journal} {\bibinfo  {journal} {Advanced Intelligent Systems}\ }\textbf {\bibinfo {volume} {5}},\ \bibinfo {pages} {2200329} (\bibinfo {year} {2023})}\BibitemShut {NoStop}%
\bibitem [{\citenamefont {Zhuo}\ \emph {et~al.}(2020)\citenamefont {Zhuo}, \citenamefont {Zhao}, \citenamefont {Xie}, \citenamefont {Hao}, \citenamefont {Xu}, \citenamefont {Zhao}, \citenamefont {Li}, \citenamefont {Knubben}, \citenamefont {Wen}, \citenamefont {Jiang} \emph {et~al.}}]{zhuo_complex_2020}%
  \BibitemOpen
  \bibfield  {author} {\bibinfo {author} {\bibfnamefont {S.}~\bibnamefont {Zhuo}}, \bibinfo {author} {\bibfnamefont {Z.}~\bibnamefont {Zhao}}, \bibinfo {author} {\bibfnamefont {Z.}~\bibnamefont {Xie}}, \bibinfo {author} {\bibfnamefont {Y.}~\bibnamefont {Hao}}, \bibinfo {author} {\bibfnamefont {Y.}~\bibnamefont {Xu}}, \bibinfo {author} {\bibfnamefont {T.}~\bibnamefont {Zhao}}, \bibinfo {author} {\bibfnamefont {H.}~\bibnamefont {Li}}, \bibinfo {author} {\bibfnamefont {E.~M.}\ \bibnamefont {Knubben}}, \bibinfo {author} {\bibfnamefont {L.}~\bibnamefont {Wen}}, \bibinfo {author} {\bibfnamefont {L.}~\bibnamefont {Jiang}}, \emph {et~al.},\ }\bibfield  {title} {\bibinfo {title} {Complex multiphase organohydrogels with programmable mechanics toward adaptive soft-matter machines},\ }\href@noop {} {\bibfield  {journal} {\bibinfo  {journal} {Science advances}\ }\textbf {\bibinfo {volume} {6}},\ \bibinfo {pages} {eaax1464} (\bibinfo {year} {2020})}\BibitemShut {NoStop}%
\bibitem [{\citenamefont {Ze}\ \emph {et~al.}(2020)\citenamefont {Ze}, \citenamefont {Kuang}, \citenamefont {Wu}, \citenamefont {Wong}, \citenamefont {Montgomery}, \citenamefont {Zhang}, \citenamefont {Kovitz}, \citenamefont {Yang}, \citenamefont {Qi},\ and\ \citenamefont {Zhao}}]{ze_magnetic_2020}%
  \BibitemOpen
  \bibfield  {author} {\bibinfo {author} {\bibfnamefont {Q.}~\bibnamefont {Ze}}, \bibinfo {author} {\bibfnamefont {X.}~\bibnamefont {Kuang}}, \bibinfo {author} {\bibfnamefont {S.}~\bibnamefont {Wu}}, \bibinfo {author} {\bibfnamefont {J.}~\bibnamefont {Wong}}, \bibinfo {author} {\bibfnamefont {S.~M.}\ \bibnamefont {Montgomery}}, \bibinfo {author} {\bibfnamefont {R.}~\bibnamefont {Zhang}}, \bibinfo {author} {\bibfnamefont {J.~M.}\ \bibnamefont {Kovitz}}, \bibinfo {author} {\bibfnamefont {F.}~\bibnamefont {Yang}}, \bibinfo {author} {\bibfnamefont {H.~J.}\ \bibnamefont {Qi}},\ and\ \bibinfo {author} {\bibfnamefont {R.}~\bibnamefont {Zhao}},\ }\bibfield  {title} {\bibinfo {title} {Magnetic shape memory polymers with integrated multifunctional shape manipulation},\ }\href@noop {} {\bibfield  {journal} {\bibinfo  {journal} {Advanced Materials}\ }\textbf {\bibinfo {volume} {32}},\ \bibinfo {pages} {1906657} (\bibinfo {year} {2020})}\BibitemShut {NoStop}%
\bibitem [{\citenamefont {Wang}\ \emph {et~al.}(2018)\citenamefont {Wang}, \citenamefont {Sim}, \citenamefont {Chen}, \citenamefont {Kim}, \citenamefont {Rao}, \citenamefont {Li}, \citenamefont {Chen}, \citenamefont {Song}, \citenamefont {Verduzco},\ and\ \citenamefont {Yu}}]{wang_soft_2018}%
  \BibitemOpen
  \bibfield  {author} {\bibinfo {author} {\bibfnamefont {C.}~\bibnamefont {Wang}}, \bibinfo {author} {\bibfnamefont {K.}~\bibnamefont {Sim}}, \bibinfo {author} {\bibfnamefont {J.}~\bibnamefont {Chen}}, \bibinfo {author} {\bibfnamefont {H.}~\bibnamefont {Kim}}, \bibinfo {author} {\bibfnamefont {Z.}~\bibnamefont {Rao}}, \bibinfo {author} {\bibfnamefont {Y.}~\bibnamefont {Li}}, \bibinfo {author} {\bibfnamefont {W.}~\bibnamefont {Chen}}, \bibinfo {author} {\bibfnamefont {J.}~\bibnamefont {Song}}, \bibinfo {author} {\bibfnamefont {R.}~\bibnamefont {Verduzco}},\ and\ \bibinfo {author} {\bibfnamefont {C.}~\bibnamefont {Yu}},\ }\bibfield  {title} {\bibinfo {title} {Soft ultrathin electronics innervated adaptive fully soft robots},\ }\href@noop {} {\bibfield  {journal} {\bibinfo  {journal} {Advanced Materials}\ }\textbf {\bibinfo {volume} {30}},\ \bibinfo {pages} {1706695} (\bibinfo {year} {2018})}\BibitemShut {NoStop}%
\bibitem [{\citenamefont {Zhang}\ \emph {et~al.}(2019)\citenamefont {Zhang}, \citenamefont {Li}, \citenamefont {Yu}, \citenamefont {Chai}, \citenamefont {Li}, \citenamefont {Wu},\ and\ \citenamefont {Jiang}}]{zhang_magnetic_2019}%
  \BibitemOpen
  \bibfield  {author} {\bibinfo {author} {\bibfnamefont {Z.}~\bibnamefont {Zhang}}, \bibinfo {author} {\bibfnamefont {X.}~\bibnamefont {Li}}, \bibinfo {author} {\bibfnamefont {X.}~\bibnamefont {Yu}}, \bibinfo {author} {\bibfnamefont {H.}~\bibnamefont {Chai}}, \bibinfo {author} {\bibfnamefont {Y.}~\bibnamefont {Li}}, \bibinfo {author} {\bibfnamefont {H.}~\bibnamefont {Wu}},\ and\ \bibinfo {author} {\bibfnamefont {S.}~\bibnamefont {Jiang}},\ }\bibfield  {title} {\bibinfo {title} {Magnetic actuation bionic robotic gripper with bistable morphing structure},\ }\href@noop {} {\bibfield  {journal} {\bibinfo  {journal} {Composite Structures}\ }\textbf {\bibinfo {volume} {229}},\ \bibinfo {pages} {111422} (\bibinfo {year} {2019})}\BibitemShut {NoStop}%
\bibitem [{\citenamefont {Alapan}\ \emph {et~al.}(2020)\citenamefont {Alapan}, \citenamefont {Karacakol}, \citenamefont {Guzelhan}, \citenamefont {Isik},\ and\ \citenamefont {Sitti}}]{alapan_reprogrammable_2020}%
  \BibitemOpen
  \bibfield  {author} {\bibinfo {author} {\bibfnamefont {Y.}~\bibnamefont {Alapan}}, \bibinfo {author} {\bibfnamefont {A.~C.}\ \bibnamefont {Karacakol}}, \bibinfo {author} {\bibfnamefont {S.~N.}\ \bibnamefont {Guzelhan}}, \bibinfo {author} {\bibfnamefont {I.}~\bibnamefont {Isik}},\ and\ \bibinfo {author} {\bibfnamefont {M.}~\bibnamefont {Sitti}},\ }\bibfield  {title} {\bibinfo {title} {Reprogrammable shape morphing of magnetic soft machines},\ }\href@noop {} {\bibfield  {journal} {\bibinfo  {journal} {Science advances}\ }\textbf {\bibinfo {volume} {6}},\ \bibinfo {pages} {eabc6414} (\bibinfo {year} {2020})}\BibitemShut {NoStop}%
\bibitem [{\citenamefont {Cheng}\ \emph {et~al.}(2018)\citenamefont {Cheng}, \citenamefont {Ren}, \citenamefont {Yang},\ and\ \citenamefont {Wei}}]{cheng_bilayer-type_2018}%
  \BibitemOpen
  \bibfield  {author} {\bibinfo {author} {\bibfnamefont {Y.}~\bibnamefont {Cheng}}, \bibinfo {author} {\bibfnamefont {K.}~\bibnamefont {Ren}}, \bibinfo {author} {\bibfnamefont {D.}~\bibnamefont {Yang}},\ and\ \bibinfo {author} {\bibfnamefont {J.}~\bibnamefont {Wei}},\ }\bibfield  {title} {\bibinfo {title} {Bilayer-type fluorescence hydrogels with intelligent response serve as temperature/ph driven soft actuators},\ }\href@noop {} {\bibfield  {journal} {\bibinfo  {journal} {Sensors and Actuators B: Chemical}\ }\textbf {\bibinfo {volume} {255}},\ \bibinfo {pages} {3117} (\bibinfo {year} {2018})}\BibitemShut {NoStop}%
\bibitem [{\citenamefont {Breger}\ \emph {et~al.}(2015)\citenamefont {Breger}, \citenamefont {Yoon}, \citenamefont {Xiao}, \citenamefont {Kwag}, \citenamefont {Wang}, \citenamefont {Fisher}, \citenamefont {Nguyen},\ and\ \citenamefont {Gracias}}]{breger_self-folding_2015}%
  \BibitemOpen
  \bibfield  {author} {\bibinfo {author} {\bibfnamefont {J.~C.}\ \bibnamefont {Breger}}, \bibinfo {author} {\bibfnamefont {C.}~\bibnamefont {Yoon}}, \bibinfo {author} {\bibfnamefont {R.}~\bibnamefont {Xiao}}, \bibinfo {author} {\bibfnamefont {H.~R.}\ \bibnamefont {Kwag}}, \bibinfo {author} {\bibfnamefont {M.~O.}\ \bibnamefont {Wang}}, \bibinfo {author} {\bibfnamefont {J.~P.}\ \bibnamefont {Fisher}}, \bibinfo {author} {\bibfnamefont {T.~D.}\ \bibnamefont {Nguyen}},\ and\ \bibinfo {author} {\bibfnamefont {D.~H.}\ \bibnamefont {Gracias}},\ }\bibfield  {title} {\bibinfo {title} {Self-folding thermo-magnetically responsive soft microgrippers},\ }\href@noop {} {\bibfield  {journal} {\bibinfo  {journal} {ACS applied materials \& interfaces}\ }\textbf {\bibinfo {volume} {7}},\ \bibinfo {pages} {3398} (\bibinfo {year} {2015})}\BibitemShut {NoStop}%
\bibitem [{\citenamefont {Pilz~da Cunha}\ \emph {et~al.}(2019)\citenamefont {Pilz~da Cunha}, \citenamefont {Foelen}, \citenamefont {van Raak}, \citenamefont {Murphy}, \citenamefont {Engels}, \citenamefont {Debije},\ and\ \citenamefont {Schenning}}]{pilz_da_cunha_untethered_2019}%
  \BibitemOpen
  \bibfield  {author} {\bibinfo {author} {\bibfnamefont {M.}~\bibnamefont {Pilz~da Cunha}}, \bibinfo {author} {\bibfnamefont {Y.}~\bibnamefont {Foelen}}, \bibinfo {author} {\bibfnamefont {R.~J.}\ \bibnamefont {van Raak}}, \bibinfo {author} {\bibfnamefont {J.~N.}\ \bibnamefont {Murphy}}, \bibinfo {author} {\bibfnamefont {T.~A.}\ \bibnamefont {Engels}}, \bibinfo {author} {\bibfnamefont {M.~G.}\ \bibnamefont {Debije}},\ and\ \bibinfo {author} {\bibfnamefont {A.~P.}\ \bibnamefont {Schenning}},\ }\bibfield  {title} {\bibinfo {title} {An untethered magnetic-and light-responsive rotary gripper: shedding light on photoresponsive liquid crystal actuators},\ }\href@noop {} {\bibfield  {journal} {\bibinfo  {journal} {Advanced Optical Materials}\ }\textbf {\bibinfo {volume} {7}},\ \bibinfo {pages} {1801643} (\bibinfo {year} {2019})}\BibitemShut {NoStop}%
\bibitem [{\citenamefont {Pan}\ \emph {et~al.}(2021)\citenamefont {Pan}, \citenamefont {Grossiord}, \citenamefont {Sol}, \citenamefont {Debije},\ and\ \citenamefont {Schenning}}]{pan_3d_2021}%
  \BibitemOpen
  \bibfield  {author} {\bibinfo {author} {\bibfnamefont {X.}~\bibnamefont {Pan}}, \bibinfo {author} {\bibfnamefont {N.}~\bibnamefont {Grossiord}}, \bibinfo {author} {\bibfnamefont {J.~A.}\ \bibnamefont {Sol}}, \bibinfo {author} {\bibfnamefont {M.~G.}\ \bibnamefont {Debije}},\ and\ \bibinfo {author} {\bibfnamefont {A.~P.}\ \bibnamefont {Schenning}},\ }\bibfield  {title} {\bibinfo {title} {3d anisotropic polyethylene as light-responsive grippers and surfing divers},\ }\href@noop {} {\bibfield  {journal} {\bibinfo  {journal} {Advanced Functional Materials}\ }\textbf {\bibinfo {volume} {31}},\ \bibinfo {pages} {2100465} (\bibinfo {year} {2021})}\BibitemShut {NoStop}%
\bibitem [{\citenamefont {Ilievski}\ \emph {et~al.}(2011)\citenamefont {Ilievski}, \citenamefont {Mazzeo}, \citenamefont {Shepherd}, \citenamefont {Chen},\ and\ \citenamefont {Whitesides}}]{ilievski_soft_2011}%
  \BibitemOpen
  \bibfield  {author} {\bibinfo {author} {\bibfnamefont {F.}~\bibnamefont {Ilievski}}, \bibinfo {author} {\bibfnamefont {A.~D.}\ \bibnamefont {Mazzeo}}, \bibinfo {author} {\bibfnamefont {R.~F.}\ \bibnamefont {Shepherd}}, \bibinfo {author} {\bibfnamefont {X.}~\bibnamefont {Chen}},\ and\ \bibinfo {author} {\bibfnamefont {G.~M.}\ \bibnamefont {Whitesides}},\ }\bibfield  {title} {\bibinfo {title} {Soft robotics for chemists},\ }\href@noop {} {\bibfield  {journal} {\bibinfo  {journal} {Angewandte Chemie}\ }\textbf {\bibinfo {volume} {123}},\ \bibinfo {pages} {1930} (\bibinfo {year} {2011})}\BibitemShut {NoStop}%
\bibitem [{\citenamefont {Li}\ \emph {et~al.}(2017)\citenamefont {Li}, \citenamefont {Vogt}, \citenamefont {Rus},\ and\ \citenamefont {Wood}}]{li_fluid-driven_2017}%
  \BibitemOpen
  \bibfield  {author} {\bibinfo {author} {\bibfnamefont {S.}~\bibnamefont {Li}}, \bibinfo {author} {\bibfnamefont {D.~M.}\ \bibnamefont {Vogt}}, \bibinfo {author} {\bibfnamefont {D.}~\bibnamefont {Rus}},\ and\ \bibinfo {author} {\bibfnamefont {R.~J.}\ \bibnamefont {Wood}},\ }\bibfield  {title} {\bibinfo {title} {Fluid-driven origami-inspired artificial muscles},\ }\href@noop {} {\bibfield  {journal} {\bibinfo  {journal} {Proceedings of the National academy of Sciences}\ }\textbf {\bibinfo {volume} {114}},\ \bibinfo {pages} {13132} (\bibinfo {year} {2017})}\BibitemShut {NoStop}%
\bibitem [{\citenamefont {Deimel}\ and\ \citenamefont {Brock}(2016)}]{deimel_novel_2016}%
  \BibitemOpen
  \bibfield  {author} {\bibinfo {author} {\bibfnamefont {R.}~\bibnamefont {Deimel}}\ and\ \bibinfo {author} {\bibfnamefont {O.}~\bibnamefont {Brock}},\ }\bibfield  {title} {\bibinfo {title} {A novel type of compliant and underactuated robotic hand for dexterous grasping},\ }\href@noop {} {\bibfield  {journal} {\bibinfo  {journal} {The International Journal of Robotics Research}\ }\textbf {\bibinfo {volume} {35}},\ \bibinfo {pages} {161} (\bibinfo {year} {2016})}\BibitemShut {NoStop}%
\bibitem [{\citenamefont {Pal}\ \emph {et~al.}(2021)\citenamefont {Pal}, \citenamefont {Restrepo}, \citenamefont {Goswami},\ and\ \citenamefont {Martinez}}]{pal2021exploiting}%
  \BibitemOpen
  \bibfield  {author} {\bibinfo {author} {\bibfnamefont {A.}~\bibnamefont {Pal}}, \bibinfo {author} {\bibfnamefont {V.}~\bibnamefont {Restrepo}}, \bibinfo {author} {\bibfnamefont {D.}~\bibnamefont {Goswami}},\ and\ \bibinfo {author} {\bibfnamefont {R.~V.}\ \bibnamefont {Martinez}},\ }\bibfield  {title} {\bibinfo {title} {Exploiting mechanical instabilities in soft robotics: Control, sensing, and actuation},\ }\href@noop {} {\bibfield  {journal} {\bibinfo  {journal} {Advanced Materials}\ }\textbf {\bibinfo {volume} {33}},\ \bibinfo {pages} {2006939} (\bibinfo {year} {2021})}\BibitemShut {NoStop}%
\bibitem [{\citenamefont {Yang}\ \emph {et~al.}(2015)\citenamefont {Yang}, \citenamefont {Mosadegh}, \citenamefont {Ainla}, \citenamefont {Lee}, \citenamefont {Khashai}, \citenamefont {Suo}, \citenamefont {Bertoldi},\ and\ \citenamefont {Whitesides}}]{yang_buckling_2015}%
  \BibitemOpen
  \bibfield  {author} {\bibinfo {author} {\bibfnamefont {D.}~\bibnamefont {Yang}}, \bibinfo {author} {\bibfnamefont {B.}~\bibnamefont {Mosadegh}}, \bibinfo {author} {\bibfnamefont {A.}~\bibnamefont {Ainla}}, \bibinfo {author} {\bibfnamefont {B.~C.~G.}\ \bibnamefont {Lee}}, \bibinfo {author} {\bibfnamefont {F.}~\bibnamefont {Khashai}}, \bibinfo {author} {\bibfnamefont {Z.}~\bibnamefont {Suo}}, \bibinfo {author} {\bibfnamefont {K.}~\bibnamefont {Bertoldi}},\ and\ \bibinfo {author} {\bibfnamefont {G.~M.}\ \bibnamefont {Whitesides}},\ }\bibfield  {title} {\bibinfo {title} {Buckling of elastomeric beams enables actuation of soft machines},\ }\href@noop {} {\bibfield  {journal} {\bibinfo  {journal} {Adv. Mater.}\ } (\bibinfo {year} {2015})}\BibitemShut {NoStop}%
\bibitem [{\citenamefont {Tang}\ \emph {et~al.}(2020)\citenamefont {Tang}, \citenamefont {Chi}, \citenamefont {Sun}, \citenamefont {Huang}, \citenamefont {Maghsoudi}, \citenamefont {Spence}, \citenamefont {Zhao}, \citenamefont {Su},\ and\ \citenamefont {Yin}}]{tang_leveraging_2020}%
  \BibitemOpen
  \bibfield  {author} {\bibinfo {author} {\bibfnamefont {Y.}~\bibnamefont {Tang}}, \bibinfo {author} {\bibfnamefont {Y.}~\bibnamefont {Chi}}, \bibinfo {author} {\bibfnamefont {J.}~\bibnamefont {Sun}}, \bibinfo {author} {\bibfnamefont {T.-H.}\ \bibnamefont {Huang}}, \bibinfo {author} {\bibfnamefont {O.~H.}\ \bibnamefont {Maghsoudi}}, \bibinfo {author} {\bibfnamefont {A.}~\bibnamefont {Spence}}, \bibinfo {author} {\bibfnamefont {J.}~\bibnamefont {Zhao}}, \bibinfo {author} {\bibfnamefont {H.}~\bibnamefont {Su}},\ and\ \bibinfo {author} {\bibfnamefont {J.}~\bibnamefont {Yin}},\ }\bibfield  {title} {\bibinfo {title} {Leveraging elastic instabilities for amplified performance: Spine-inspired high-speed and high-force soft robots},\ }\href@noop {} {\bibfield  {journal} {\bibinfo  {journal} {Science advances}\ }\textbf {\bibinfo {volume} {6}},\ \bibinfo {pages} {eaaz6912} (\bibinfo {year} {2020})}\BibitemShut {NoStop}%
\bibitem [{\citenamefont {Lin}\ \emph {et~al.}(2021)\citenamefont {Lin}, \citenamefont {Zhang}, \citenamefont {Tang}, \citenamefont {Jiao}, \citenamefont {Wang}, \citenamefont {Wang}, \citenamefont {Zhong}, \citenamefont {Zhu}, \citenamefont {Hu}, \citenamefont {Yang} \emph {et~al.}}]{lin_bioinspired_2021}%
  \BibitemOpen
  \bibfield  {author} {\bibinfo {author} {\bibfnamefont {Y.}~\bibnamefont {Lin}}, \bibinfo {author} {\bibfnamefont {C.}~\bibnamefont {Zhang}}, \bibinfo {author} {\bibfnamefont {W.}~\bibnamefont {Tang}}, \bibinfo {author} {\bibfnamefont {Z.}~\bibnamefont {Jiao}}, \bibinfo {author} {\bibfnamefont {J.}~\bibnamefont {Wang}}, \bibinfo {author} {\bibfnamefont {W.}~\bibnamefont {Wang}}, \bibinfo {author} {\bibfnamefont {Y.}~\bibnamefont {Zhong}}, \bibinfo {author} {\bibfnamefont {P.}~\bibnamefont {Zhu}}, \bibinfo {author} {\bibfnamefont {Y.}~\bibnamefont {Hu}}, \bibinfo {author} {\bibfnamefont {H.}~\bibnamefont {Yang}}, \emph {et~al.},\ }\bibfield  {title} {\bibinfo {title} {A bioinspired stress-response strategy for high-speed soft grippers},\ }\href@noop {} {\bibfield  {journal} {\bibinfo  {journal} {Advanced Science}\ }\textbf {\bibinfo {volume} {8}},\ \bibinfo {pages} {2102539} (\bibinfo {year} {2021})}\BibitemShut {NoStop}%
\bibitem [{\citenamefont {Qi}\ \emph {et~al.}(2022)\citenamefont {Qi}, \citenamefont {Xiang}, \citenamefont {Ho},\ and\ \citenamefont {Rossiter}}]{qi_sea-anemone-inspired_2022}%
  \BibitemOpen
  \bibfield  {author} {\bibinfo {author} {\bibfnamefont {Q.}~\bibnamefont {Qi}}, \bibinfo {author} {\bibfnamefont {C.}~\bibnamefont {Xiang}}, \bibinfo {author} {\bibfnamefont {V.~A.}\ \bibnamefont {Ho}},\ and\ \bibinfo {author} {\bibfnamefont {J.}~\bibnamefont {Rossiter}},\ }\bibfield  {title} {\bibinfo {title} {A sea-anemone-inspired, multifunctional, bistable gripper},\ }\href@noop {} {\bibfield  {journal} {\bibinfo  {journal} {Soft Robotics}\ }\textbf {\bibinfo {volume} {9}},\ \bibinfo {pages} {1040} (\bibinfo {year} {2022})}\BibitemShut {NoStop}%
\bibitem [{\citenamefont {Jin}\ \emph {et~al.}(2023)\citenamefont {Jin}, \citenamefont {Yang}, \citenamefont {Maldonado}, \citenamefont {Lee}, \citenamefont {Figueroa}, \citenamefont {Full},\ and\ \citenamefont {Yang}}]{jin_ultrafast_2023}%
  \BibitemOpen
  \bibfield  {author} {\bibinfo {author} {\bibfnamefont {L.}~\bibnamefont {Jin}}, \bibinfo {author} {\bibfnamefont {Y.}~\bibnamefont {Yang}}, \bibinfo {author} {\bibfnamefont {B.~O.~T.}\ \bibnamefont {Maldonado}}, \bibinfo {author} {\bibfnamefont {S.~D.}\ \bibnamefont {Lee}}, \bibinfo {author} {\bibfnamefont {N.}~\bibnamefont {Figueroa}}, \bibinfo {author} {\bibfnamefont {R.~J.}\ \bibnamefont {Full}},\ and\ \bibinfo {author} {\bibfnamefont {S.}~\bibnamefont {Yang}},\ }\bibfield  {title} {\bibinfo {title} {Ultrafast, programmable, and electronics-free soft robots enabled by snapping metacaps},\ }\href@noop {} {\bibfield  {journal} {\bibinfo  {journal} {Advanced Intelligent Systems}\ ,\ \bibinfo {pages} {2300039}} (\bibinfo {year} {2023})}\BibitemShut {NoStop}%
\bibitem [{\citenamefont {Lidmar}\ \emph {et~al.}(2003)\citenamefont {Lidmar}, \citenamefont {Mirny},\ and\ \citenamefont {Nelson}}]{lidmar_virus_2003}%
  \BibitemOpen
  \bibfield  {author} {\bibinfo {author} {\bibfnamefont {J.}~\bibnamefont {Lidmar}}, \bibinfo {author} {\bibfnamefont {L.}~\bibnamefont {Mirny}},\ and\ \bibinfo {author} {\bibfnamefont {D.~R.}\ \bibnamefont {Nelson}},\ }\bibfield  {title} {\bibinfo {title} {Virus shapes and buckling transitions in spherical shells},\ }\href@noop {} {\bibfield  {journal} {\bibinfo  {journal} {Physical Review E}\ }\textbf {\bibinfo {volume} {68}},\ \bibinfo {pages} {051910} (\bibinfo {year} {2003})}\BibitemShut {NoStop}%
\bibitem [{\citenamefont {Glaenzer}\ \emph {et~al.}(2017)\citenamefont {Glaenzer}, \citenamefont {Peter}, \citenamefont {Thomas},\ and\ \citenamefont {Hagelueken}}]{glaenzer_peldor_2017}%
  \BibitemOpen
  \bibfield  {author} {\bibinfo {author} {\bibfnamefont {J.}~\bibnamefont {Glaenzer}}, \bibinfo {author} {\bibfnamefont {M.~F.}\ \bibnamefont {Peter}}, \bibinfo {author} {\bibfnamefont {G.~H.}\ \bibnamefont {Thomas}},\ and\ \bibinfo {author} {\bibfnamefont {G.}~\bibnamefont {Hagelueken}},\ }\bibfield  {title} {\bibinfo {title} {Peldor spectroscopy reveals two defined states of a sialic acid trap transporter sbp in solution},\ }\href@noop {} {\bibfield  {journal} {\bibinfo  {journal} {Biophysical Journal}\ }\textbf {\bibinfo {volume} {112}},\ \bibinfo {pages} {109} (\bibinfo {year} {2017})}\BibitemShut {NoStop}%
\bibitem [{\citenamefont {Schaechter}(2002)}]{schaechter2002stroll}%
  \BibitemOpen
  \bibfield  {author} {\bibinfo {author} {\bibfnamefont {E.}~\bibnamefont {Schaechter}},\ }\href@noop {} {\bibinfo {title} {A stroll with the moulds}} (\bibinfo {year} {2002})\BibitemShut {NoStop}%
\bibitem [{\citenamefont {Katifori}\ \emph {et~al.}(2010)\citenamefont {Katifori}, \citenamefont {Alben}, \citenamefont {Cerda}, \citenamefont {Nelson},\ and\ \citenamefont {Dumais}}]{katifori_foldable_2010}%
  \BibitemOpen
  \bibfield  {author} {\bibinfo {author} {\bibfnamefont {E.}~\bibnamefont {Katifori}}, \bibinfo {author} {\bibfnamefont {S.}~\bibnamefont {Alben}}, \bibinfo {author} {\bibfnamefont {E.}~\bibnamefont {Cerda}}, \bibinfo {author} {\bibfnamefont {D.~R.}\ \bibnamefont {Nelson}},\ and\ \bibinfo {author} {\bibfnamefont {J.}~\bibnamefont {Dumais}},\ }\bibfield  {title} {\bibinfo {title} {Foldable structures and the natural design of pollen grains},\ }\href@noop {} {\bibfield  {journal} {\bibinfo  {journal} {Proceedings of the National Academy of Sciences}\ }\textbf {\bibinfo {volume} {107}},\ \bibinfo {pages} {7635} (\bibinfo {year} {2010})}\BibitemShut {NoStop}%
\bibitem [{\citenamefont {Forterre}\ \emph {et~al.}(2005)\citenamefont {Forterre}, \citenamefont {Skotheim}, \citenamefont {Dumais},\ and\ \citenamefont {Mahadevan}}]{forterre_how_2005}%
  \BibitemOpen
  \bibfield  {author} {\bibinfo {author} {\bibfnamefont {Y.}~\bibnamefont {Forterre}}, \bibinfo {author} {\bibfnamefont {J.~M.}\ \bibnamefont {Skotheim}}, \bibinfo {author} {\bibfnamefont {J.}~\bibnamefont {Dumais}},\ and\ \bibinfo {author} {\bibfnamefont {L.}~\bibnamefont {Mahadevan}},\ }\bibfield  {title} {\bibinfo {title} {How the venus flytrap snaps},\ }\href@noop {} {\bibfield  {journal} {\bibinfo  {journal} {Nature}\ }\textbf {\bibinfo {volume} {433}},\ \bibinfo {pages} {421} (\bibinfo {year} {2005})}\BibitemShut {NoStop}%
\bibitem [{\citenamefont {Nakane}\ and\ \citenamefont {Miyata}(2007)}]{nakane_cytoskeletal_2007}%
  \BibitemOpen
  \bibfield  {author} {\bibinfo {author} {\bibfnamefont {D.}~\bibnamefont {Nakane}}\ and\ \bibinfo {author} {\bibfnamefont {M.}~\bibnamefont {Miyata}},\ }\bibfield  {title} {\bibinfo {title} {Cytoskeletal “jellyfish” structure of mycoplasma mobile},\ }\href@noop {} {\bibfield  {journal} {\bibinfo  {journal} {Proceedings of the National Academy of Sciences}\ }\textbf {\bibinfo {volume} {104}},\ \bibinfo {pages} {19518} (\bibinfo {year} {2007})}\BibitemShut {NoStop}%
\bibitem [{\citenamefont {Tauzin}\ \emph {et~al.}(2014)\citenamefont {Tauzin}, \citenamefont {Starnes}, \citenamefont {Becker}, \citenamefont {Lam},\ and\ \citenamefont {Huttenlocher}}]{tauzin_redox_2014}%
  \BibitemOpen
  \bibfield  {author} {\bibinfo {author} {\bibfnamefont {S.}~\bibnamefont {Tauzin}}, \bibinfo {author} {\bibfnamefont {T.~W.}\ \bibnamefont {Starnes}}, \bibinfo {author} {\bibfnamefont {F.~B.}\ \bibnamefont {Becker}}, \bibinfo {author} {\bibfnamefont {P.-y.}\ \bibnamefont {Lam}},\ and\ \bibinfo {author} {\bibfnamefont {A.}~\bibnamefont {Huttenlocher}},\ }\bibfield  {title} {\bibinfo {title} {Redox and src family kinase signaling control leukocyte wound attraction and neutrophil reverse migration},\ }\href@noop {} {\bibfield  {journal} {\bibinfo  {journal} {Journal of Cell Biology}\ }\textbf {\bibinfo {volume} {207}},\ \bibinfo {pages} {589} (\bibinfo {year} {2014})}\BibitemShut {NoStop}%
\bibitem [{\citenamefont {Burreson}(2020)}]{burreson_marine_2020}%
  \BibitemOpen
  \bibfield  {author} {\bibinfo {author} {\bibfnamefont {E.~M.}\ \bibnamefont {Burreson}},\ }\bibfield  {title} {\bibinfo {title} {Marine and estuarine leeches (hirudinida: Ozobranchidae and piscicolidae) of australia and new zealand with a key to the species},\ }\href@noop {} {\bibfield  {journal} {\bibinfo  {journal} {Invertebrate systematics}\ }\textbf {\bibinfo {volume} {34}},\ \bibinfo {pages} {235} (\bibinfo {year} {2020})}\BibitemShut {NoStop}%
\bibitem [{\citenamefont {Tramacere}\ \emph {et~al.}(2014)\citenamefont {Tramacere}, \citenamefont {Kovalev}, \citenamefont {Kleinteich}, \citenamefont {Gorb},\ and\ \citenamefont {Mazzolai}}]{tramacere_structure_2014}%
  \BibitemOpen
  \bibfield  {author} {\bibinfo {author} {\bibfnamefont {F.}~\bibnamefont {Tramacere}}, \bibinfo {author} {\bibfnamefont {A.}~\bibnamefont {Kovalev}}, \bibinfo {author} {\bibfnamefont {T.}~\bibnamefont {Kleinteich}}, \bibinfo {author} {\bibfnamefont {S.~N.}\ \bibnamefont {Gorb}},\ and\ \bibinfo {author} {\bibfnamefont {B.}~\bibnamefont {Mazzolai}},\ }\bibfield  {title} {\bibinfo {title} {Structure and mechanical properties of octopus vulgaris suckers},\ }\href@noop {} {\bibfield  {journal} {\bibinfo  {journal} {Journal of The Royal Society Interface}\ }\textbf {\bibinfo {volume} {11}},\ \bibinfo {pages} {20130816} (\bibinfo {year} {2014})}\BibitemShut {NoStop}%
\bibitem [{\citenamefont {Costello}\ \emph {et~al.}(2021)\citenamefont {Costello}, \citenamefont {Colin}, \citenamefont {Dabiri}, \citenamefont {Gemmell}, \citenamefont {Lucas},\ and\ \citenamefont {Sutherland}}]{costello_hydrodynamics_2021}%
  \BibitemOpen
  \bibfield  {author} {\bibinfo {author} {\bibfnamefont {J.~H.}\ \bibnamefont {Costello}}, \bibinfo {author} {\bibfnamefont {S.~P.}\ \bibnamefont {Colin}}, \bibinfo {author} {\bibfnamefont {J.~O.}\ \bibnamefont {Dabiri}}, \bibinfo {author} {\bibfnamefont {B.~J.}\ \bibnamefont {Gemmell}}, \bibinfo {author} {\bibfnamefont {K.~N.}\ \bibnamefont {Lucas}},\ and\ \bibinfo {author} {\bibfnamefont {K.~R.}\ \bibnamefont {Sutherland}},\ }\bibfield  {title} {\bibinfo {title} {The hydrodynamics of jellyfish swimming},\ }\href@noop {} {\bibfield  {journal} {\bibinfo  {journal} {Annual Review of Marine Science}\ }\textbf {\bibinfo {volume} {13}},\ \bibinfo {pages} {375} (\bibinfo {year} {2021})}\BibitemShut {NoStop}%
\bibitem [{\citenamefont {Gage}\ and\ \citenamefont {Tyler}(1999)}]{gage_deep-sea_1999}%
  \BibitemOpen
  \bibfield  {author} {\bibinfo {author} {\bibfnamefont {J.~D.}\ \bibnamefont {Gage}}\ and\ \bibinfo {author} {\bibfnamefont {P.~A.}\ \bibnamefont {Tyler}},\ }\href@noop {} {\emph {\bibinfo {title} {Deep-Sea Biology}}},\ \bibinfo {edition} {reprint with corrections}\ ed.\ (\bibinfo  {publisher} {Univ. Press},\ \bibinfo {year} {1999})\BibitemShut {NoStop}%
\bibitem [{\citenamefont {Fothergill}(2001)}]{fothergill_blue_2001}%
  \BibitemOpen
  \bibfield  {author} {\bibinfo {author} {\bibfnamefont {A.}~\bibnamefont {Fothergill}},\ }\href@noop {} {\bibinfo {title} {Blue planet: The deep}} (\bibinfo {year} {2001})\BibitemShut {NoStop}%
\bibitem [{\citenamefont {Zoelly}(1915)}]{zoelly_ueber_1915}%
  \BibitemOpen
  \bibfield  {author} {\bibinfo {author} {\bibfnamefont {R.}~\bibnamefont {Zoelly}},\ }\href@noop {} {\bibinfo {title} {Ueber ein knickungsproblem an der kugelschale}} (\bibinfo {year} {1915})\BibitemShut {NoStop}%
\bibitem [{\citenamefont {Timoshenko}\ and\ \citenamefont {Gere}(2009)}]{timoshenko_theory_2009}%
  \BibitemOpen
  \bibfield  {author} {\bibinfo {author} {\bibfnamefont {S.}~\bibnamefont {Timoshenko}}\ and\ \bibinfo {author} {\bibfnamefont {J.~M.}\ \bibnamefont {Gere}},\ }\href@noop {} {\emph {\bibinfo {title} {Theory of elastic stability}}},\ \bibinfo {edition} {2nd}\ ed.\ (\bibinfo  {publisher} {Dover Publications},\ \bibinfo {year} {2009})\ \bibinfo {note} {{OCLC}: ocn294885242}\BibitemShut {NoStop}%
\bibitem [{\citenamefont {Gent}(1996)}]{gent1996new}%
  \BibitemOpen
  \bibfield  {author} {\bibinfo {author} {\bibfnamefont {A.~N.}\ \bibnamefont {Gent}},\ }\bibfield  {title} {\bibinfo {title} {A new constitutive relation for rubber},\ }\href@noop {} {\bibfield  {journal} {\bibinfo  {journal} {Rubber chemistry and technology}\ }\textbf {\bibinfo {volume} {69}},\ \bibinfo {pages} {59} (\bibinfo {year} {1996})}\BibitemShut {NoStop}%
\bibitem [{\citenamefont {Lee}\ \emph {et~al.}(2016{\natexlab{a}})\citenamefont {Lee}, \citenamefont {L{\'o}pez~Jim{\'e}nez}, \citenamefont {Marthelot}, \citenamefont {Hutchinson},\ and\ \citenamefont {Reis}}]{lee2016geometric}%
  \BibitemOpen
  \bibfield  {author} {\bibinfo {author} {\bibfnamefont {A.}~\bibnamefont {Lee}}, \bibinfo {author} {\bibfnamefont {F.}~\bibnamefont {L{\'o}pez~Jim{\'e}nez}}, \bibinfo {author} {\bibfnamefont {J.}~\bibnamefont {Marthelot}}, \bibinfo {author} {\bibfnamefont {J.~W.}\ \bibnamefont {Hutchinson}},\ and\ \bibinfo {author} {\bibfnamefont {P.~M.}\ \bibnamefont {Reis}},\ }\bibfield  {title} {\bibinfo {title} {The geometric role of precisely engineered imperfections on the critical buckling load of spherical elastic shells},\ }\href@noop {} {\bibfield  {journal} {\bibinfo  {journal} {Journal of Applied Mechanics}\ }\textbf {\bibinfo {volume} {83}},\ \bibinfo {pages} {111005} (\bibinfo {year} {2016}{\natexlab{a}})}\BibitemShut {NoStop}%
\bibitem [{\citenamefont {Hutchinson}\ and\ \citenamefont {Thompson}(2017)}]{hutchinson2017nonlinear}%
  \BibitemOpen
  \bibfield  {author} {\bibinfo {author} {\bibfnamefont {J.~W.}\ \bibnamefont {Hutchinson}}\ and\ \bibinfo {author} {\bibfnamefont {J.~M.~T.}\ \bibnamefont {Thompson}},\ }\bibfield  {title} {\bibinfo {title} {Nonlinear buckling behaviour of spherical shells: barriers and symmetry-breaking dimples},\ }\href@noop {} {\bibfield  {journal} {\bibinfo  {journal} {Philosophical Transactions of the Royal Society A: Mathematical, Physical and Engineering Sciences}\ }\textbf {\bibinfo {volume} {375}},\ \bibinfo {pages} {20160154} (\bibinfo {year} {2017})}\BibitemShut {NoStop}%
\bibitem [{\citenamefont {Lee}\ \emph {et~al.}(2016{\natexlab{b}})\citenamefont {Lee}, \citenamefont {Brun}, \citenamefont {Marthelot}, \citenamefont {Balestra}, \citenamefont {Gallaire},\ and\ \citenamefont {Reis}}]{lee2016fabrication}%
  \BibitemOpen
  \bibfield  {author} {\bibinfo {author} {\bibfnamefont {A.}~\bibnamefont {Lee}}, \bibinfo {author} {\bibfnamefont {P.-T.}\ \bibnamefont {Brun}}, \bibinfo {author} {\bibfnamefont {J.}~\bibnamefont {Marthelot}}, \bibinfo {author} {\bibfnamefont {G.}~\bibnamefont {Balestra}}, \bibinfo {author} {\bibfnamefont {F.}~\bibnamefont {Gallaire}},\ and\ \bibinfo {author} {\bibfnamefont {P.~M.}\ \bibnamefont {Reis}},\ }\bibfield  {title} {\bibinfo {title} {Fabrication of slender elastic shells by the coating of curved surfaces},\ }\href@noop {} {\bibfield  {journal} {\bibinfo  {journal} {Nature communications}\ }\textbf {\bibinfo {volume} {7}},\ \bibinfo {pages} {11155} (\bibinfo {year} {2016}{\natexlab{b}})}\BibitemShut {NoStop}%
\bibitem [{\citenamefont {Lee}\ \emph {et~al.}(2021)\citenamefont {Lee}, \citenamefont {Park},\ and\ \citenamefont {Holmes}}]{lee2021stimuli}%
  \BibitemOpen
  \bibfield  {author} {\bibinfo {author} {\bibfnamefont {J.-H.}\ \bibnamefont {Lee}}, \bibinfo {author} {\bibfnamefont {H.~S.}\ \bibnamefont {Park}},\ and\ \bibinfo {author} {\bibfnamefont {D.~P.}\ \bibnamefont {Holmes}},\ }\bibfield  {title} {\bibinfo {title} {Stimuli-responsive shell theory},\ }\href@noop {} {\bibfield  {journal} {\bibinfo  {journal} {Mathematics and Mechanics of Solids}\ ,\ \bibinfo {pages} {10812865231159676}} (\bibinfo {year} {2021})}\BibitemShut {NoStop}%
\bibitem [{\citenamefont {Jampani}\ \emph {et~al.}(2019)\citenamefont {Jampani}, \citenamefont {Volpe}, \citenamefont {Reguengo~de Sousa}, \citenamefont {Ferreira~Machado}, \citenamefont {Yakacki},\ and\ \citenamefont {Lagerwall}}]{jampani2019liquid}%
  \BibitemOpen
  \bibfield  {author} {\bibinfo {author} {\bibfnamefont {V.}~\bibnamefont {Jampani}}, \bibinfo {author} {\bibfnamefont {R.}~\bibnamefont {Volpe}}, \bibinfo {author} {\bibfnamefont {K.}~\bibnamefont {Reguengo~de Sousa}}, \bibinfo {author} {\bibfnamefont {J.}~\bibnamefont {Ferreira~Machado}}, \bibinfo {author} {\bibfnamefont {C.}~\bibnamefont {Yakacki}},\ and\ \bibinfo {author} {\bibfnamefont {J.}~\bibnamefont {Lagerwall}},\ }\bibfield  {title} {\bibinfo {title} {Liquid crystal elastomer shell actuators with negative order parameter},\ }\href@noop {} {\bibfield  {journal} {\bibinfo  {journal} {Science advances}\ }\textbf {\bibinfo {volume} {5}},\ \bibinfo {pages} {eaaw2476} (\bibinfo {year} {2019})}\BibitemShut {NoStop}%
\bibitem [{\citenamefont {Yan}\ \emph {et~al.}(2021)\citenamefont {Yan}, \citenamefont {Pezzulla}, \citenamefont {Cruveiller}, \citenamefont {Abbasi},\ and\ \citenamefont {Reis}}]{yan2021magneto}%
  \BibitemOpen
  \bibfield  {author} {\bibinfo {author} {\bibfnamefont {D.}~\bibnamefont {Yan}}, \bibinfo {author} {\bibfnamefont {M.}~\bibnamefont {Pezzulla}}, \bibinfo {author} {\bibfnamefont {L.}~\bibnamefont {Cruveiller}}, \bibinfo {author} {\bibfnamefont {A.}~\bibnamefont {Abbasi}},\ and\ \bibinfo {author} {\bibfnamefont {P.~M.}\ \bibnamefont {Reis}},\ }\bibfield  {title} {\bibinfo {title} {Magneto-active elastic shells with tunable buckling strength},\ }\href@noop {} {\bibfield  {journal} {\bibinfo  {journal} {Nature communications}\ }\textbf {\bibinfo {volume} {12}},\ \bibinfo {pages} {2831} (\bibinfo {year} {2021})}\BibitemShut {NoStop}%
\bibitem [{\citenamefont {Ogden}(1997)}]{ogden1997non}%
  \BibitemOpen
  \bibfield  {author} {\bibinfo {author} {\bibfnamefont {R.~W.}\ \bibnamefont {Ogden}},\ }\href@noop {} {\emph {\bibinfo {title} {Non-linear elastic deformations}}}\ (\bibinfo  {publisher} {Courier Corporation},\ \bibinfo {year} {1997})\BibitemShut {NoStop}%
\bibitem [{\citenamefont {Vasios}\ \emph {et~al.}(2020)\citenamefont {Vasios}, \citenamefont {Gross}, \citenamefont {Soifer}, \citenamefont {Overvelde},\ and\ \citenamefont {Bertoldi}}]{vasios2020harnessing}%
  \BibitemOpen
  \bibfield  {author} {\bibinfo {author} {\bibfnamefont {N.}~\bibnamefont {Vasios}}, \bibinfo {author} {\bibfnamefont {A.~J.}\ \bibnamefont {Gross}}, \bibinfo {author} {\bibfnamefont {S.}~\bibnamefont {Soifer}}, \bibinfo {author} {\bibfnamefont {J.~T.}\ \bibnamefont {Overvelde}},\ and\ \bibinfo {author} {\bibfnamefont {K.}~\bibnamefont {Bertoldi}},\ }\bibfield  {title} {\bibinfo {title} {Harnessing viscous flow to simplify the actuation of fluidic soft robots},\ }\href@noop {} {\bibfield  {journal} {\bibinfo  {journal} {Soft robotics}\ }\textbf {\bibinfo {volume} {7}},\ \bibinfo {pages} {1} (\bibinfo {year} {2020})}\BibitemShut {NoStop}%
\bibitem [{\citenamefont {Gorissen}\ \emph {et~al.}(2020)\citenamefont {Gorissen}, \citenamefont {Melancon}, \citenamefont {Vasios}, \citenamefont {Torbati},\ and\ \citenamefont {Bertoldi}}]{gorissen_inflatable_2020}%
  \BibitemOpen
  \bibfield  {author} {\bibinfo {author} {\bibfnamefont {B.}~\bibnamefont {Gorissen}}, \bibinfo {author} {\bibfnamefont {D.}~\bibnamefont {Melancon}}, \bibinfo {author} {\bibfnamefont {N.}~\bibnamefont {Vasios}}, \bibinfo {author} {\bibfnamefont {M.}~\bibnamefont {Torbati}},\ and\ \bibinfo {author} {\bibfnamefont {K.}~\bibnamefont {Bertoldi}},\ }\bibfield  {title} {\bibinfo {title} {Inflatable soft jumper inspired by shell snapping},\ }\href@noop {} {\bibfield  {journal} {\bibinfo  {journal} {Science Robotics}\ }\textbf {\bibinfo {volume} {5}},\ \bibinfo {pages} {eabb1967} (\bibinfo {year} {2020})}\BibitemShut {NoStop}%
\end{thebibliography}%

\clearpage
\section{Supplementary Material}
\subsection{Section S1. Finite Element Simulations}

\begin{figure*}
    \centering
    \includegraphics[width=\textwidth,keepaspectratio]{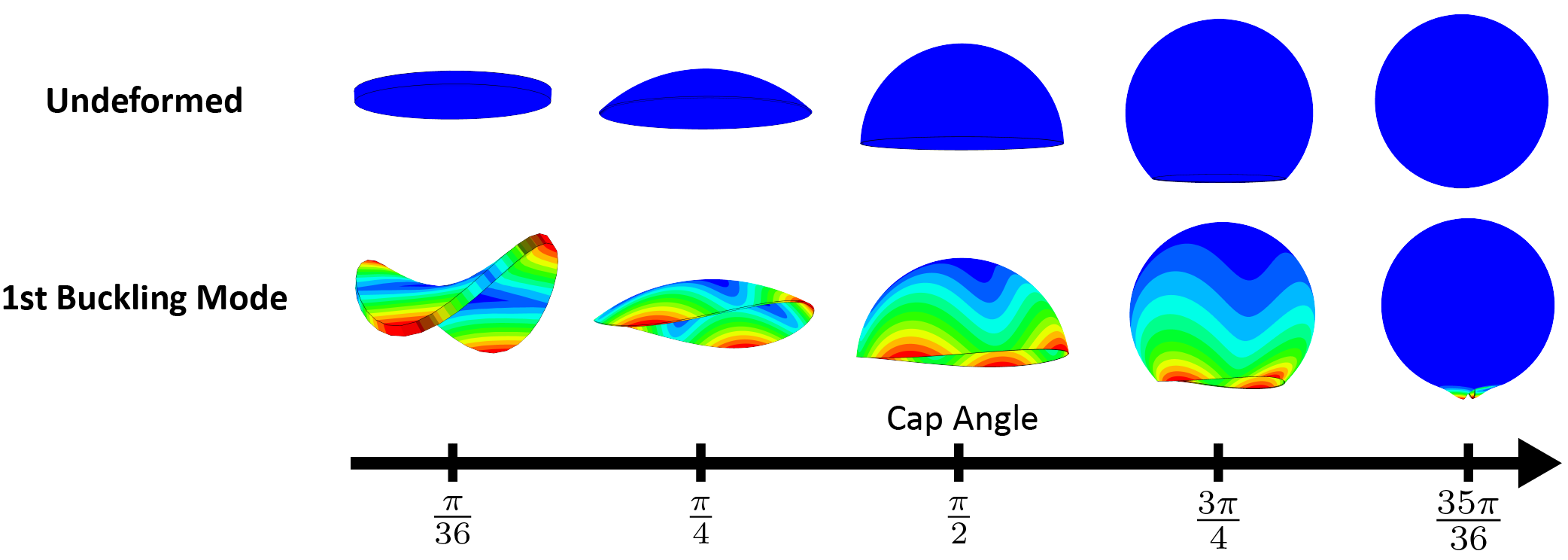}
    \caption{\textbf{FE analyses of free hemispheres.} Undeformed and buckling mode shape of free hemispheres for different cap angles 
$\theta$.}
    \label{suppfig:capangle}
\end{figure*}

To gather a better understanding of the buckling behavior of free thin-shell domes, we conduct finite element (FE) analyses using the commercial package ABAQUS 2020/Standard.
We create the hemispherical shell models using a structured mesh with 4-node quadrilateral shell elements (ABAQUS element type: S4R).
We model the hemispherical shell, fabricated out of Zhermack Elite Double 32, using a nearly incompressible neo-Hookean \cite{ogden1997non} ~constitutive model with shear modulus $G = 0.375 \text{ MPa}$ and Poisson's ratio $\nu = 0.4998$ \cite{vasios2020harnessing}. 
Then, we apply a fixed boundary condition at the hemisphere pole and impose a uniform pressure load to the inner surface of the hemisphere within a buckling step FE analysis.
We utilize a meshing seed sufficiently small that successive refinements produce a deviation in eigenvalue under $2\%$.
Then, we run buckling analyses to extract the first five eigenmodes of the hemisphere for multiple cap angles $\theta$ and shell thicknesses $h$.
Though varying the cap angle $\theta$ significantly modifies the shell structure, the mode shape of interest can be observed across the different geometrical configurations by tracking the deformation of the lobes along the major and minor axes of the boundary edge as shown in Fig. \ref{suppfig:capangle}.

To represent the soft gripper, we add a circular membrane on the equatorial plane to create a closed cavity within the hemispherical shell with cap angle $\theta = \pi/2$.
We model the film made of EcoFlex 00-30 (Smooth-On) material using an incompressible Gent material model \cite{gent1996new} ~(UHYPER user defined subroutine) with shear modulus $G = 19.43 \text{ kPa}$ and extension limit $J = 37.54$ \cite{vasios2020harnessing}. 
Similarly to the hemispherical shell, we utilize the S4R 4-node shell elements to mesh the film.
To model the behavior of the soft gripper, we introduce a geometric imperfection obtained from the buckling analysis conducted on the hemispherical shell.
To avoid biasing results towards a specific mode shape, we extract imperfections from the buckling analysis of both the free and clamped hemispherical shell configurations.
To mimic the experiments conducted controlling the volume of the fluid cavity, we numerically simulate the change in fluid volume by fictitiously varying the temperature of the fluid cavity according to the following relationship

\begin{equation}
    \frac{\Delta V}{V_0}=3\alpha_{\text{T}}\Delta T
    \label{suppeq:fluidtemp}
\end{equation}

where $\Delta V$ is the volume variation, $V_0$ is the initial volume, $\alpha_{\text{T}}$ is the coefficient of thermal expansion, and $\Delta T$ is the temperature variation.
We assume the hydraulic fluid with a bulk modulus $B = 2000 \text{ MPa}$, fluid density $\rho = 1000 \text{ kg}/\text{m}^3$, and coefficient of thermal expansion $\alpha_{\text{T}} = 1 \text{ m}/(\text{m}\cdot\text{K})$ \cite{gorissen_inflatable_2020}. 
We run the simulations by progressively varying the temperature up to the corresponding  volume change $\Delta V = 84\% V_0$ of the fluid cavity.

\begin{figure*}
    \centering
    \includegraphics[keepaspectratio]{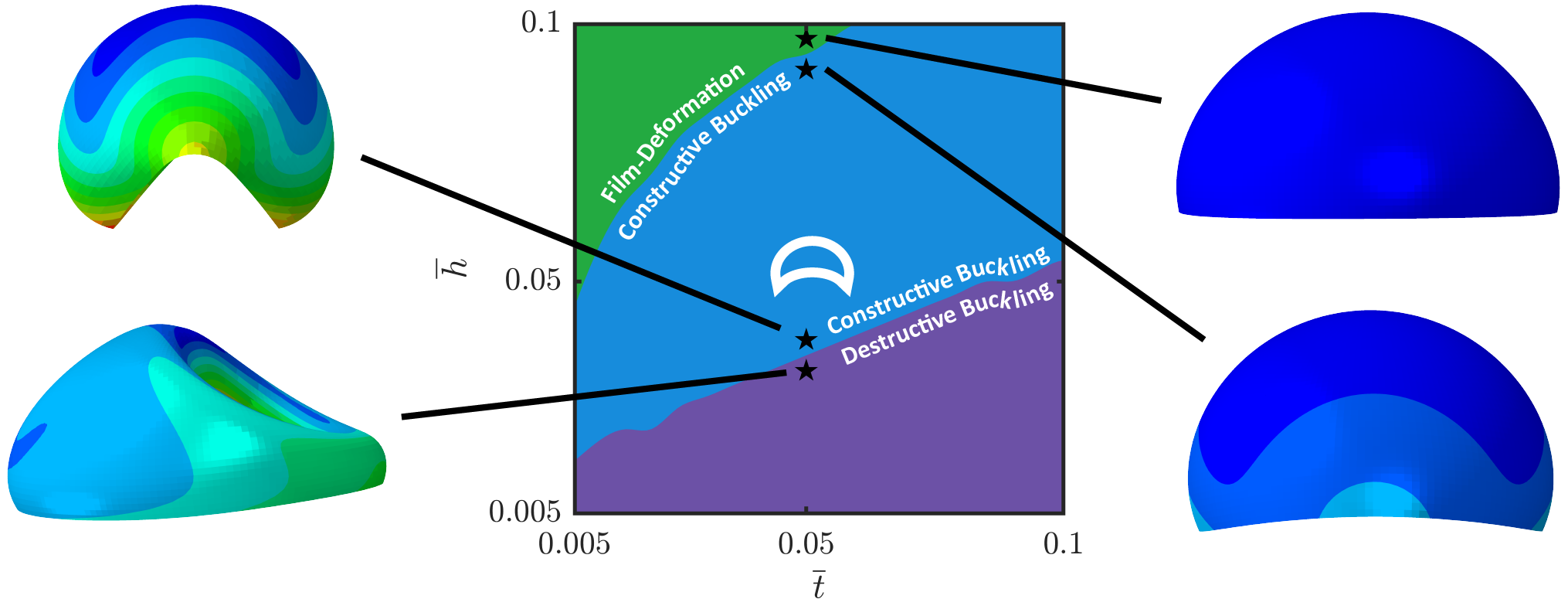}
    \caption{\textbf{Numerical Design Space}. Identification of the boundary regions between the three regimes that characterize the response of the soft hemispherical actuator as a function of the slenderness ratio $\bar{h}$ and normalized film thickness $\bar{t}$, respectively.}
    \label{suppfig:dsbreakdown}
\end{figure*}

\begin{figure*}
    \centering
    \includegraphics[scale=0.9]{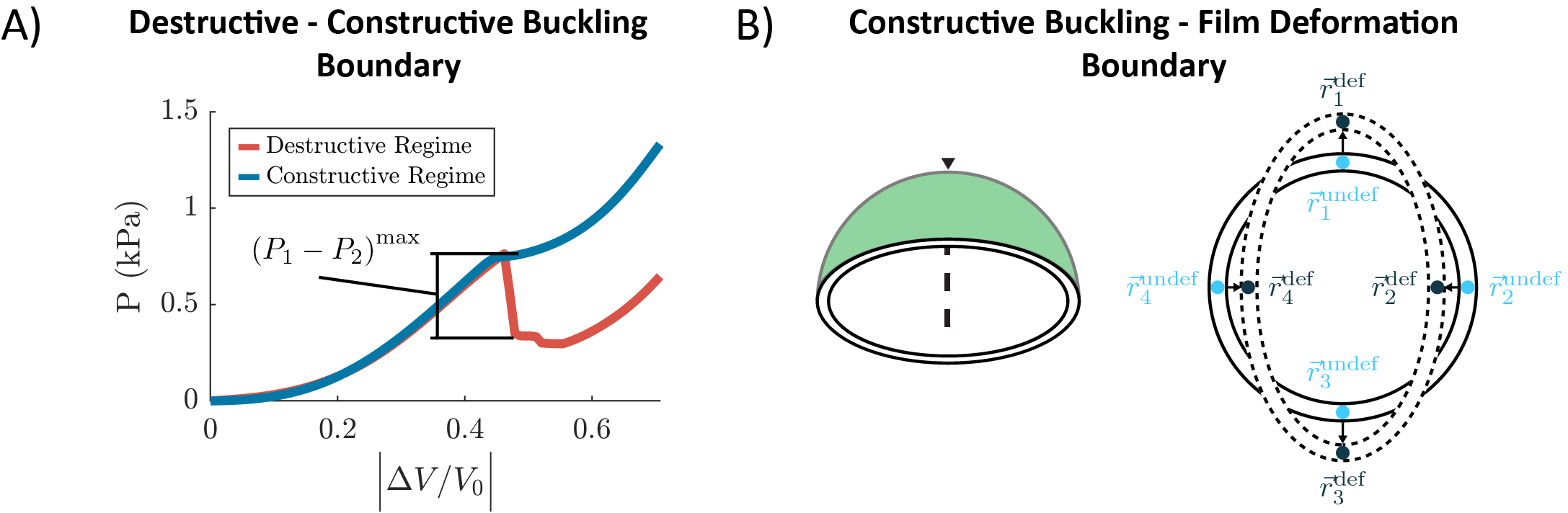}
    \caption{\textbf{Design Space Metrics.} \textbf{A)} Metrics to distinguish between the `destructive' (Regime 1) and `constructive' (Regime 2) buckling regime,  \textbf{B)} Metrics to distinguish between the `constructive' buckling regime (Regime 2) and the film deformation regime (Regime 3).} 
    \label{suppfig:metrics}
\end{figure*}

We explore the design space by simulating the deflation of the soft gripper using the dynamic implicit solver for ranges of slenderness ratio $0.005<\bar{h}<0.1$ and normalized film thickness $0.005<\bar{t}<0.1$ (Fig. \ref{suppfig:dsbreakdown}) for a total of 320 simulations. In these simulations, for both rubber materials, we consider a density $\rho = 1070 \text{ kg}/\text{m}^3$ and a stiffness proportional Rayleigh damping $\beta = 0.01$.
To guarantee quasi-static conditions by checking the kinetic energy, we set the time period of the dynamic implicit analysis to $20 \text{ s}$, with a minimum time increment of $1e-4 \text{ s}$, maximum time increment of $0.2 \text{ s}$, and $100,000,000$ maximum number of increments.
We automatically conduct the classification of the three regimes - the `destructive' buckling regime (Regime 1),  the `constructive' buckling regime (Regime 2), and  the film deformation regime (Regime 3) - by evaluating predetermined metrics based on the pressure-volume curves and on the soft gripper deformation (Fig. \ref{suppfig:metrics}).

In the `destructive' buckling case, the buckling point manifests a sharp drop in pressure, whereas in the  `constructive' buckling case  a slope change can be observed at the buckling onset as shown in Fig. \ref{suppfig:metrics}A.
As such, we introduce a metric $M_{\text{DC}}$ based on the pressure difference between consecutive pressure readings given by

\begin{equation}
    M_{\text{DC}}=\left|\left(P_1-P_2\right)^{\text{max}}/P_1\right|
    \label{suppeq:mdc}
\end{equation}

where $P_1$ and $P_2$ are pressures values obtained for two consecutive time instants and $\left(P_1-P_2\right)^{\text{max}}$ represents the largest difference between two consecutive data points during the deflation cycle.
We set a threshold of $M_{\text{DC}}=15\%$ meaning that for greater values, the sample falls within the destructive buckling regime.

Since no pressure drop appears in the film deformation regime (Regime 3), we need to introduce a different metric to distinguish between the constructive buckling and film deformation regimes.
Given that not every sample that begins deforming into the desired constructive buckling mode shape can be categorized as useful for grasping tasks, we determine the boundary between Regime 2 and 3  defining a metric $M_{\text{CF}}$ based on the displacement of the equatorial nodes along the major and minor axes of deformation (Fig. \ref{suppfig:metrics}B) given by

\begin{equation}
    M_{\text{CF}}=\left|\frac{||\vec{r}_{1}^{ \, \text{def}}-\vec{r}_{3}^{\, \text{def}}||}{||\vec{r}_{1}^{\, \text{undef}}-\vec{r}_{3}^{\, \text{undef}}||}-\frac{||\vec{r}_{2}^{\, \text{def}}-\vec{r}_{4}^{\, \text{def}}||}{||\vec{r}_{2}^{\, \text{undef}}-\vec{r}_{4}^{\, \text{undef}}||}\right|
    \label{suppeq:mcf}
\end{equation}

where $\vec{r}^{\, \text{undef}}$ and $\vec{r}^{\, \text{def}}$ refer to the undeformed and deformed positions of the nodes, respectively, as shown in Fig. \ref{suppfig:metrics}B.
We set the threshold to $M_{\text{CF}} = 10\%$ indicating that for greater values, the sample falls within the `constructive' buckling regime.

\subsection{Section S2. Manufacturing}

\begin{figure*}
    \centering
    \includegraphics[width=\textwidth,keepaspectratio]{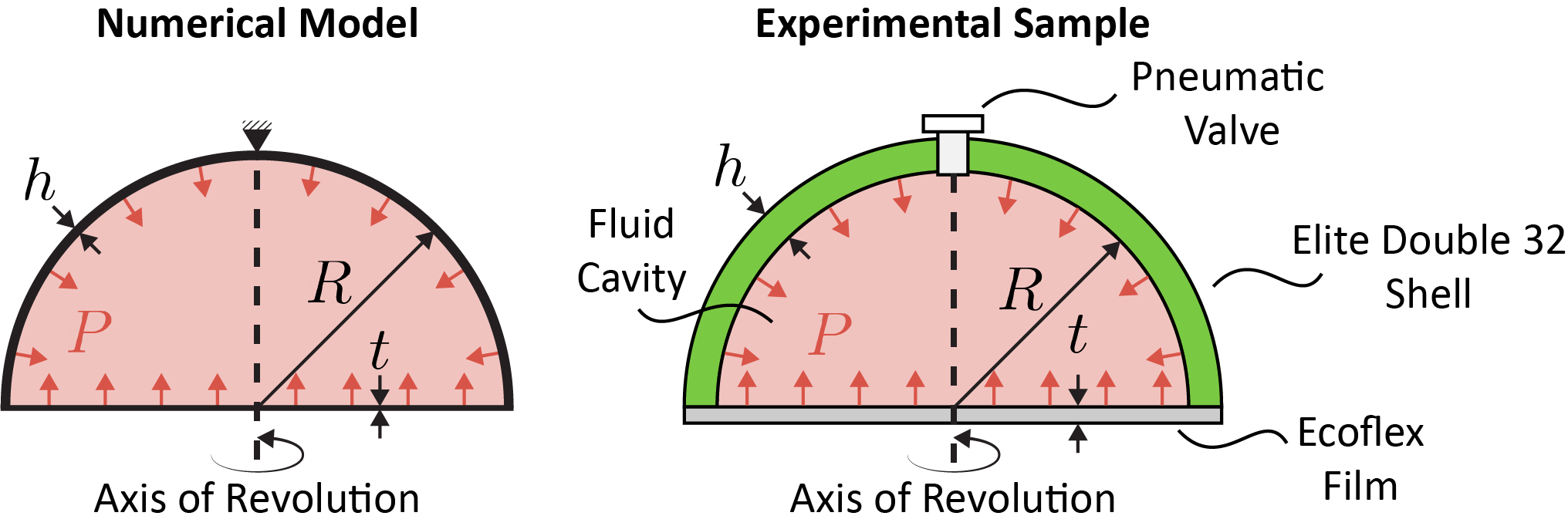}
    \caption{\textbf{``Pac-Man'' Gripper geometry.}  Schematics of the numerical model and equivalent experimental sample  of the hemispherical gripper.}
    \label{suppfig:geom}
\end{figure*}

The hemispherical gripper design is composed of a thin hemispherical shell attached to a thin film along the equatorial line to seal off an isolated fluid cavity in the structure (Fig. \ref{suppfig:geom}).
A pneumatic valve is used to enable volume changes within the fluid cavity.
For our experimental samples, we cast the hemispherical shells using Zhermack Elite Double 32, and the thin film is manufactured using Ecoflex 00-30 (Smooth-On).
Casting the experimental samples is a multi-step process based on viscous coating process for creating thin and soft shells \cite{lee2016fabrication}. 
The steps for our manufacturing process, as shown in Fig. \ref{suppfig:manufacturing}, are broken down as follows:

\begin{figure*}
    \centering
    \includegraphics[keepaspectratio]{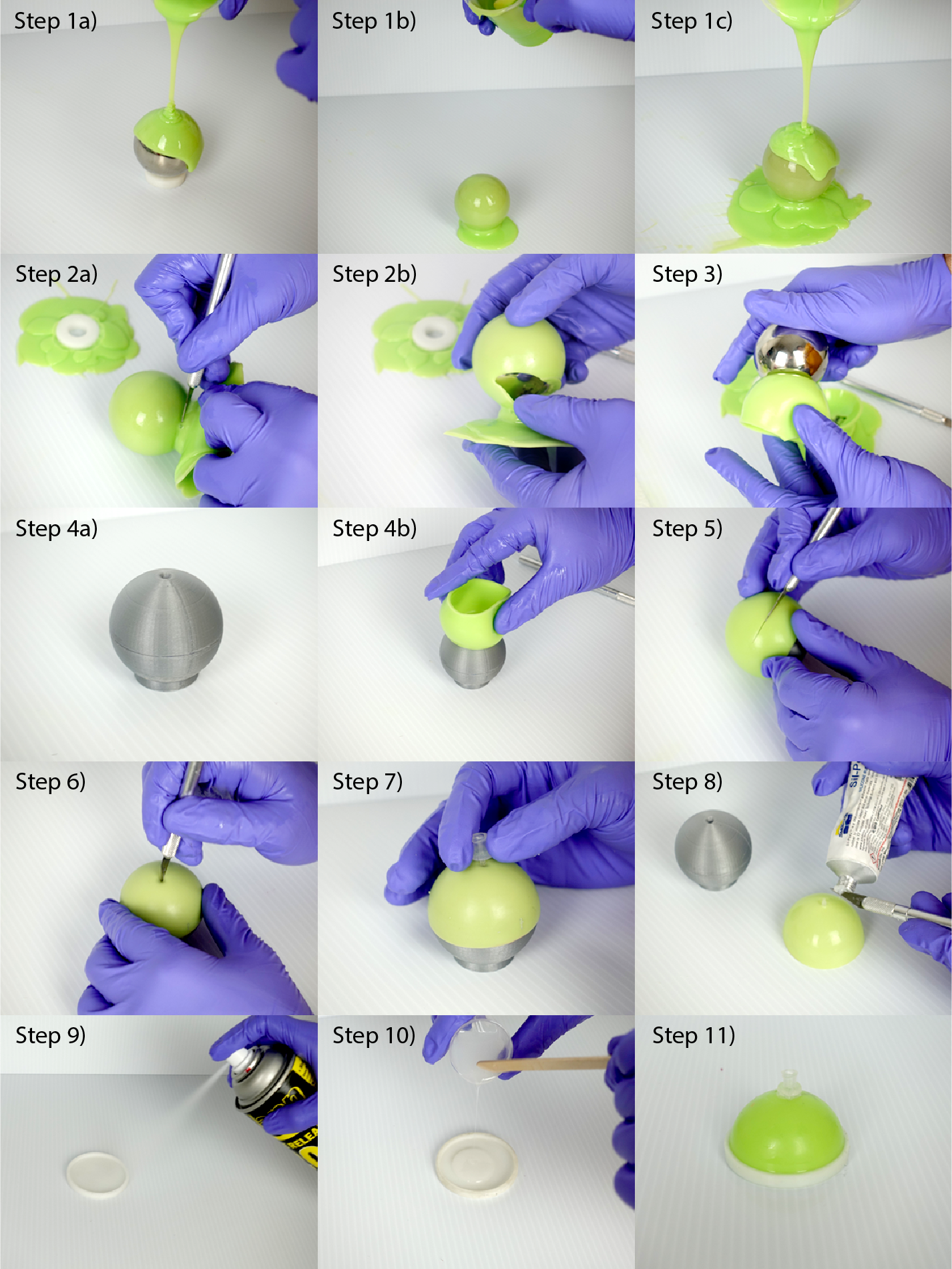}
    \caption{\textbf{Manufacturing process of the ``Pac-Man'' gripper.} Snapshots of the steps required to manufacture our hemispherical gripper.}
    \label{suppfig:manufacturing}
\end{figure*}

\begin{itemize}
    \item[] \textbf{Step 1:} The uncured polymer Elite Double 32 is mixed and let sit for a selected amount of time $\sim 300$ s. After that, it is poured onto a smooth metal sphere and the material is then allowed to cure. 
    This step can be repeated multiple times to create multiple layers to increase the thickness of the manufactured shell.
    \item[] \textbf{Step 2:} Once the final layer is cured, the shell is separated from any excess material that overflowed around the base using a sharp blade.
    \item[] \textbf{Step 3:} The shell is removed from the metal sphere by gently inverting it, mitigating the risk of the material tearing during the process.
    \item[] \textbf{Step 4:} The shell is placed onto a 3D-printed spherical piece by gently inverting the sample a second time, so that the shell returns to its initial orientation.
    \item[] \textbf{Step 5:} A sharp blade is used to cut around the equatorial line of the hemisphere, leveraging the inset groove in the 3D-printed sphere to align the cut.
    \item[] \textbf{Step 6:} The blade is used to create an incision at the top of the hemisphere to insert the pneumatic valve.
    \item[] \textbf{Step 7:} The pneumatic valve is inserted into the top of the hemisphere.
    \item[] \textbf{Step 8:} A small amount of Sil-Poxy Silicone Rubber Adhesive is applied to the bottom of the pneumatic valve to secure it to the hemispherical shell.
    \item[] \textbf{Step 9:} A release agent (Ease Release 200 spray, Mann Release Technologies) is applied to a shallow cylindrical well.
    \item[] \textbf{Step 10:} The uncured polymer Ecoflex 00-30 is mixed, poured into a cylindrical well, and allowed to disperse to cover the entire bottom of the well.
    \item[] \textbf{Step 11:} The hemispherical shell is placed within the cylindrical well and immersed in the Ecoflex 00-30 to ensure sufficient bonding area to form a robust seal while the polymer is curing to form the membrane. 
\end{itemize}

\begin{figure*}
    \centering
    \includegraphics[keepaspectratio]{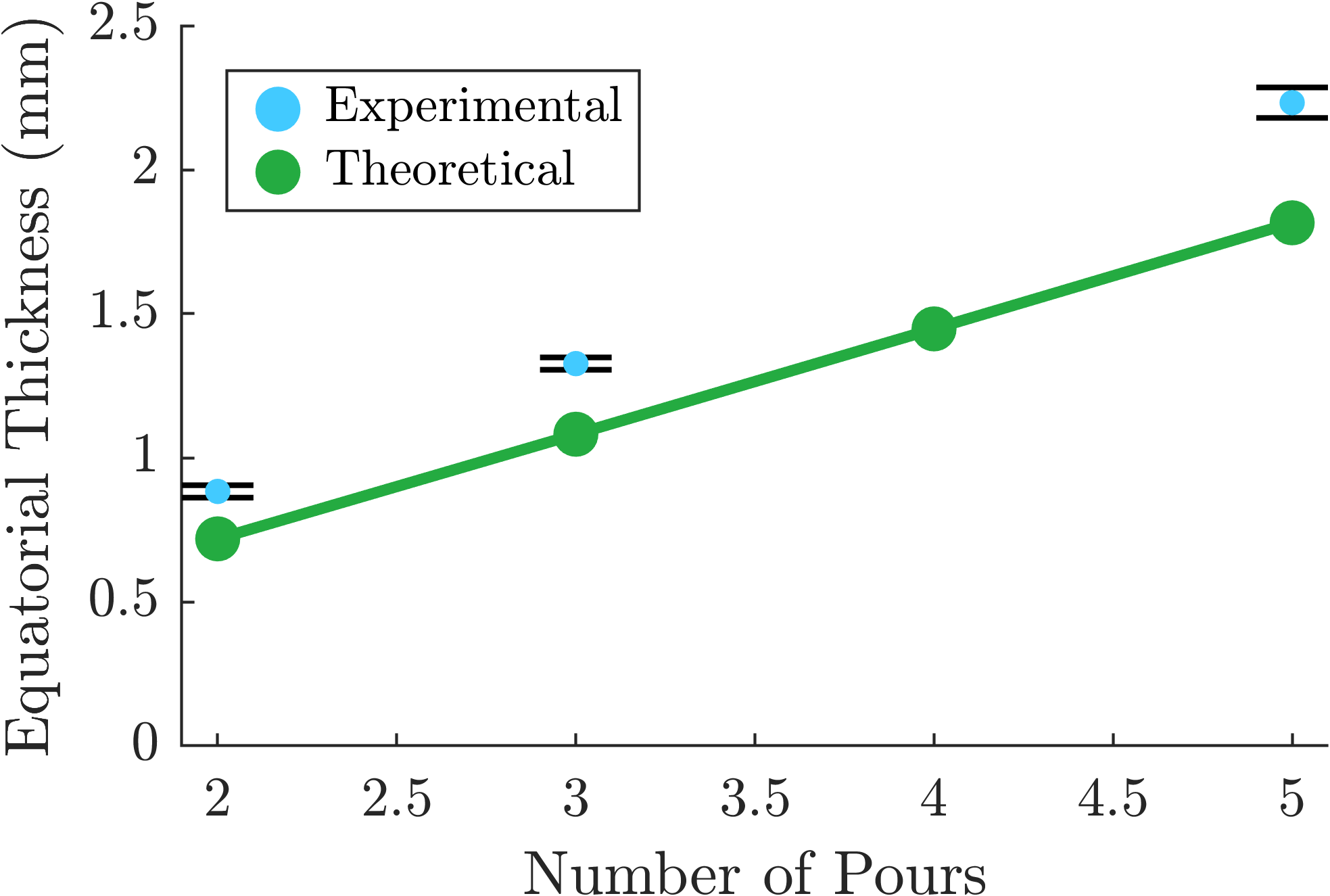}
    \caption{\textbf{Hemisphere Thickness Plot.} Analytically predicted (Eq. \ref{eq:rn}) and experimentally measured equatorial thickness of 15 hemispherical shells (five shells for each thickness). Black bars indicate the standard deviation of the measured thicknesses.}
    \label{suppfig:thickness}
\end{figure*}

We used the manufacturing process outlined here to produce all experimental samples used within this study. 

The process yields samples with controllable thicknesses, which can be demonstrated to follow the trend of the formula obtained by Lee \textit{et al.} \cite{lee2016fabrication} ~as shown in Fig. \ref{suppfig:thickness}.
In that study, the formula to determine the thickness of a thin spherical shell based on material and mixing parameters is given by \cite{lee2016fabrication} 

\begin{equation}
    h_{\text{f}} \approx \sqrt{\frac{3 \mu_0 R}{4 \rho g K}} \left(1+\frac{\phi^2}{10}\right)
\end{equation}

wherein

\begin{equation}
    K=\biggl\{\frac{k-e^{-\beta \tau_{\text{c}}}}{\beta}\biggr\}+\biggl\{\frac{\tau_{\text{c}}e^{-\beta \tau_{\text{c}}}}{\alpha-1}\biggr\}
\end{equation}

and

\begin{equation}
    k=e^{-\beta \tau_{\text{w}}}
\end{equation}

where $h_{\text{f}}$ is the hemispherical shell thickness, $\tau_{\text{c}}$ is the cure time of the material being poured, $\tau_{\text{w}}$ is the waiting time between when the material is mixed and poured, $R$ is the radius of the spherical mold, $\rho$ is the density of the material, $\mu_0$ is the initial viscosity of the fluid, $g$ is gravitational acceleration, $\phi$ is the zenith angle, and $\alpha$ and $\beta$ are fitting parameters.
For our model, we utilize the previously reported value for $\rho$, and we leverage values for $\mu_0$, $\alpha$, $\beta$, and $\tau_{\text{c}}$ reported by Lee \textit{et al.} \cite{lee2016fabrication}  for VPS-32 (namely, we utilize $\mu_0=7.1 \text{ Pa}\cdot\text{s}$, $\alpha=5.3$, $\beta=2.06 \cdot 10^{-3}$, and $\tau_{\text{c}}=574 \text{ s}$).

To manufacture our shells, we introduce an augmented version of this pour-over manufacturing approach wherein we cure multiple layers atop each other to increase the thickness of the shell beyond the range of possibility with a single pour.
Since each pour of the polymer could be approximately described by the same equation with a growing sphere radius, we describe our model with the following equation 

\begin{equation}
    R_{\text{N}}\approx R_0+\sum\limits^{\text{N}}_{\text{i}=1}\sqrt{\frac{3 \mu_0 R_{\text{i}-1}}{4 \rho g K}} \left(1+\frac{\phi^2}{10}\right)
        \label{eq:rn}
\end{equation}

\begin{equation}
    h_{\text{f}}=R_{\text{N}}-R_0
\end{equation}

where $R_0=0.025\text{ m}$ is the initial radius of the spherical mold.

In Fig. \ref{suppfig:thickness} we report the comparison between the analytically predicted and experimental collected equatorial thickness for 15 hemisphere samples (five for each thickness). A five-minutes sitting time ($\tau_{\text{w}}=300\text{ s}$) between mixing and pouring is used to cure each individual layer for our samples. 
Setting time is the variable that can be  precisely controlled to tune the shell thickness; in our experimental samples, we vary in setting time by $\pm 60 \text{ s}$ to achieve the specific thicknesses of interest.
Each of the samples uses a 50 mm diameter metal sphere as a base mold, which translates to a 50 mm inner diameter for the resulting hemispherical shells.
To replicate the three shell regimes found in the numerical simulations, we focus on manufacturing shells with three ranges of thickness $h$: (\textit{i}) `destructive' buckling (Regime 1)  $h_1 = 0.8 \text{ mm}$, (\textit{ii}) `constructive' buckling (Regime 2)  $1.1 \text{ mm} < h_2 < 1.46 \text{ mm}$, and (\textit{iii}) film deformation regime (Regime 3) $h_3 = 2.35 \text{ mm}$. In the manufactured samples, the  film thickness $t$ is $t_1 = 1.9 \text{ mm}$, $t_2 = 1.0 \text{ mm}$, and $t_3 = 0.85 \text{ mm}$, for Regime 1, 2, and 3, respectively.

\subsection{Section S3. Experimental Set-up and Testing}

\begin{figure*}
    \centering
    \includegraphics[width=\textwidth,keepaspectratio]{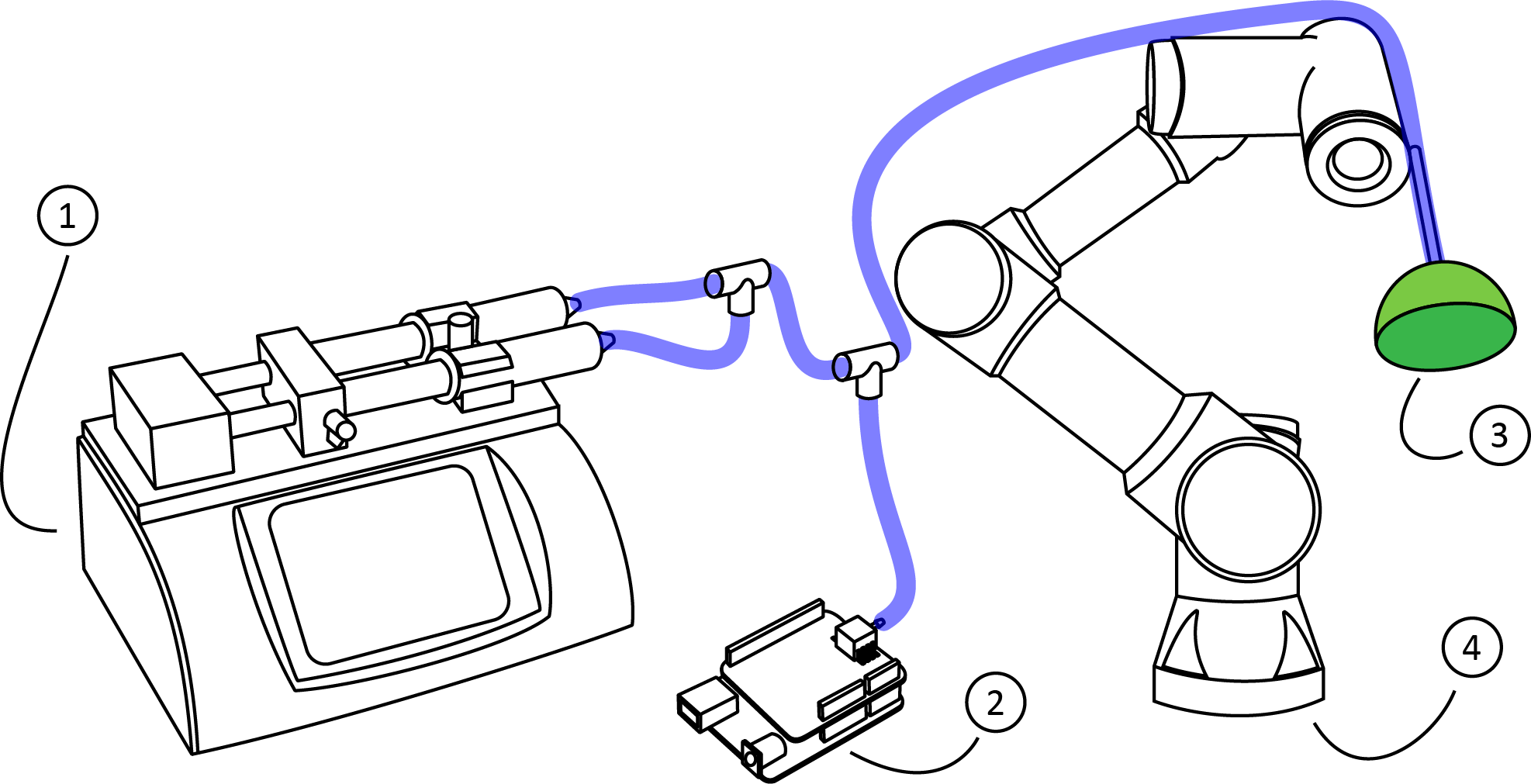}
    \caption{\textbf{Experimental setup for the grasping tests.} Schematic of the experimental setup composed by (1) a syringe pump (PHD Ultra 4400 Series Harvard Apparatus), (2) a pressure sensor shield board (FRDMSTBCDP5004 NXP), (3) the ``Pac-Man'' gripper, and (4) a robotic arm (Universal Robots UR3e).}
    \label{suppfig:experiment1}
\end{figure*}

To validate our numerical findings, we perform quasi-static inflation experiments and measure the pressure-volume curves of the soft device in all the three regimes.
The experimental set-up (Fig. \ref{suppfig:experiment1}) includes a syringe pump (PHD Ultra 4400 Series Harvard Apparatus), which pumps air out/in the hemispherical gripper at a rate of $70$ mL/min until a deflation volume of $25$ mL is reached.
This deflation target volume is set to demonstrate a full range of deformation across all three regimes.
Beyond this targeted volume, we encounter the risk of clogging the pneumatic valve from excessive deformation of the film.
The syringe pump is connected to a pressure sensor shield board (FRDMSTBCDP5004 NXP) which uses a pressure sensor (MPXV5004DP) and it is powered by an Arduino UNO micro-controller board.
The syringe pump is then connected to the hemispherical gripper through a pneumatic valve.
Finally, the hemispherical gripper is mounted to a robotic arm (Universal Robots UR3e).
A custom MATLAB script monitors readings from the Arduino UNO via serial stream, and it also interfaces with the Harvard Apparatus serial controls.
The script synchronizes collected data points with the acquisition of frames from two cameras (Sony RX100 V) oriented to provide a front-view and bottom-view of the soft gripper.

\subsubsection{A. Compressibility Effects}

Because air is used as working fluid, corrections need to be made to the raw data collected within the experiments to account for the compressibility of the fluid.
The pressure corrections are found using a simple rearrangement of Boyle's Law, which, for our experiment, is broken down in the form shown below,

\begin{equation}
    P_0^{\text{sys}}V_0^{\text{sys}}=P^{\text{sys}}V^{\text{sys}}
\end{equation}
where $P_0^{\text{sys}}$ is the initial pressure within the system, which is comprised of the hemisphere, the syringe, and the connecting tubes, $V_0^{\text{sys}}$ is the initial volume within the system, $P^{\text{sys}}$ and $V^{\text{sys}}$ are the final pressure and  volume within the system, respectively.
Since the fluid within the system is in a closed container, we expand the equation further

\begin{equation}
 V_0^{\text{sys}}=V_0^{\text{syringe}}+V_0^{\text{tubes}}+V_0^{\text{hemisphere}}
\end{equation}
where $V_0^{\text{syringe}}$, $V_0^{\text{tubes}}$, and $V_0^{\text{hemisphere}}$ represent the initial fluid volumes within the syringe, tubes, and hemisphere respectively.
The experiments investigate the deflation behavior of the hemispheres, so the sign convention on the hemisphere's volume is inverted

\begin{equation}
    V^{\text{sys}}=V_0^{\text{sys}}+\Delta V^{\text{syringe}}-\Delta V^{\text{hemisphere}}
\end{equation}

\begin{equation}
    \frac{P_0^{\text{sys}}}{P^{\text{sys}}}V_0^{\text{sys}}=V_0^{\text{sys}}+\Delta V^{\text{syringe}}-\Delta V^{\text{hemisphere}}
\end{equation}
where $\Delta V^{\text{syringe}}$ represents the final minus the initial volume within the syringe (expected to be a positive quantity for deflation, since the syringe extracts fluid from the hemisphere) and $\Delta V^{\text{hemisphere}}$ represents the initial minus the final volume within the hemisphere (which, due to the inverted sign convention, will also be positive for deflation).
The final form of the equation utilized to correct for compressibility effects is given by

\begin{equation}
    \Delta V^{\text{hemisphere}}=\Delta V^{\text{syringe}}+\frac{P^{\text{sys}}-P_0^{\text{sys}}}{P^{\text{sys}}}V_0^{\text{sys}}
    \label{suppeq:compressibility}
\end{equation}

\begin{figure*}
    \centering
    \includegraphics[keepaspectratio]{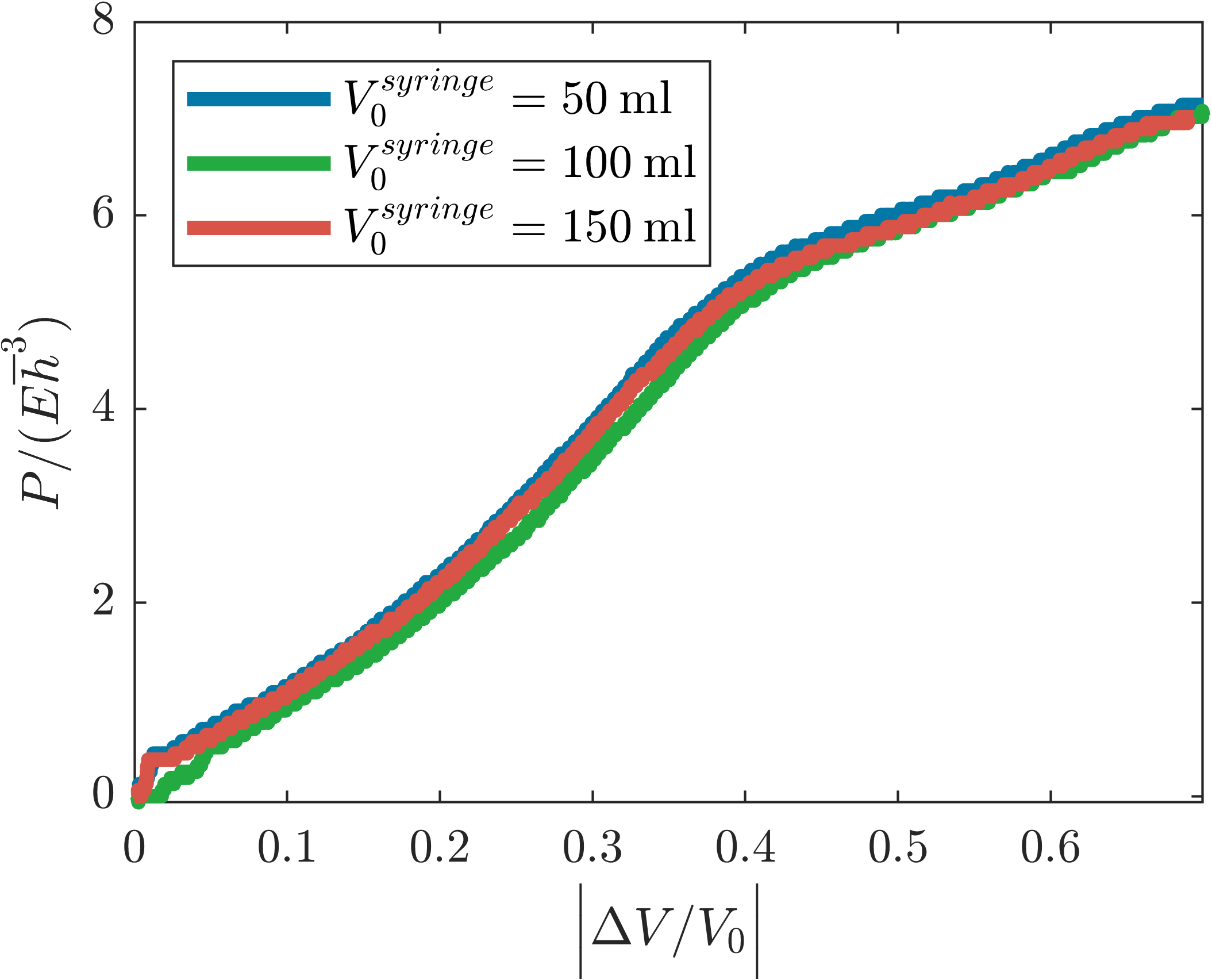}
    \caption{\textbf{Compressibility Plots.} Compressibility-corrected experimental pressure-volume data of an hemisphere ($\bar{h} = 0.052 $ and $\bar{t} = 0.048 $) tested with different initial syringe volumes $V_0^{syringe}$.}
    \label{suppfig:compressibility}
\end{figure*}

The results of three discrete deflation trials for a Regime 2 hemisphere ($\bar{h} = 0.052 $ and $\bar{t} = 0.048 $) after accounting for compressibility effects are shown in Fig. \ref{suppfig:compressibility}.
The three trials deflate at 70 mL/min with different initial volume settings within the syringe pump (50 mL, 100 mL, and 150 mL).
After accounting for compressibility using Eq. S\ref{suppeq:compressibility}, the difference in results is negligible. 

\subsubsection{B. Grasping Trials}

Grasping trials with several objects of different shape and stiffness are performed with the experimental setup shown in Fig. \ref{suppfig:experiment1}.
We use the PolyScope programming language to create customized movement profiles for the robotic arm where the soft gripper is mounted on.
A MATLAB script runs a time-based sequence of deflation and inflation steps on the syringe pump simultaneously to the movement of the robotic arm.
Before performing the grasping trials, we collect a reference pressure-volume curve of the soft gripper to characterize the `constructive' buckling behavior without the effects of any object interaction.
This reference dataset is then compared with the response of the hemisphere during each grasping trial to characterize contact, buckling, and release key points.
We used simple metrics to label each key point leveraging only pressure-volume data available live during objects' manipulation.

\begin{figure*}
    \centering
    \includegraphics[width=\textwidth,keepaspectratio]{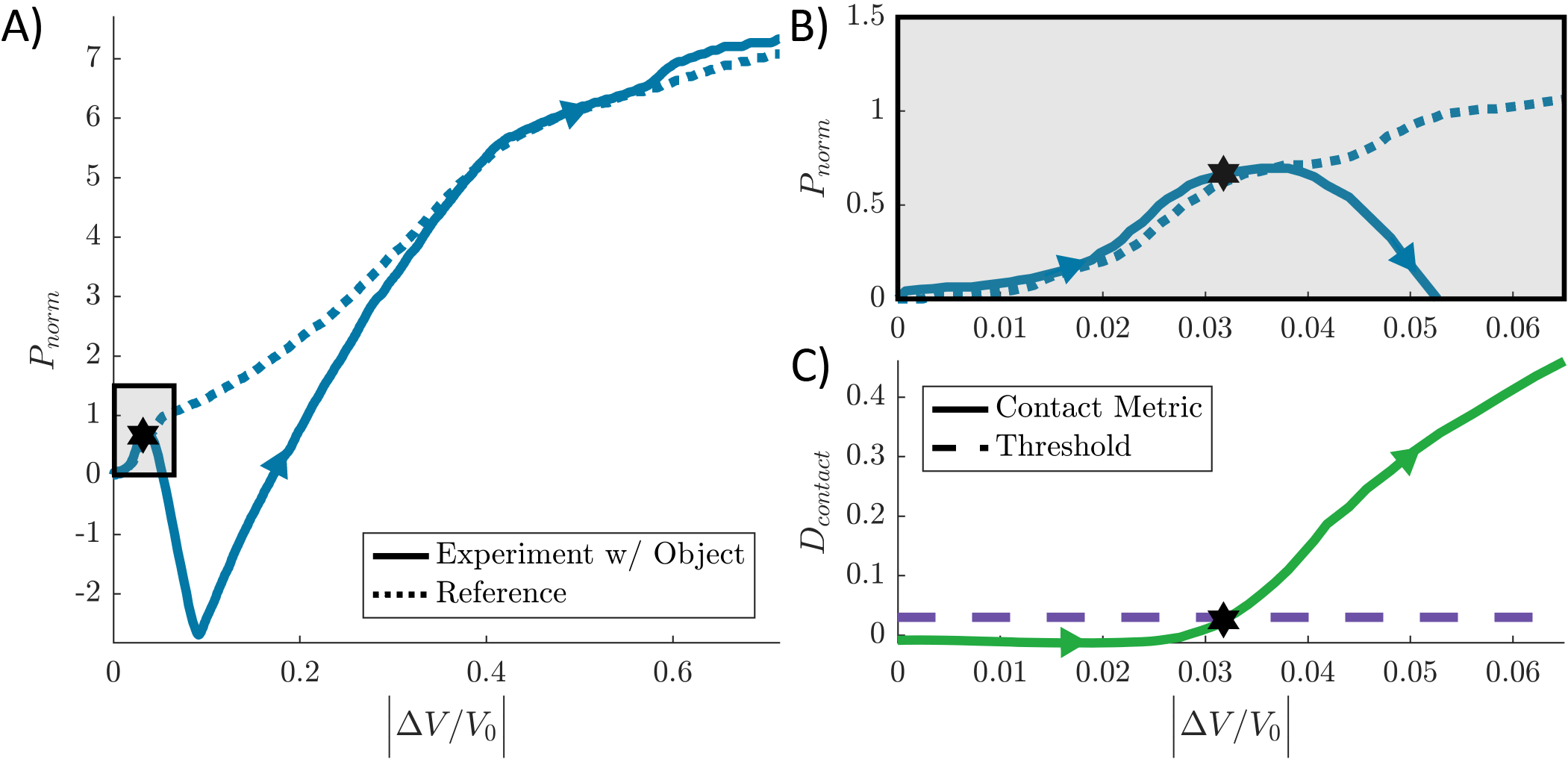}
    \caption{\textbf{Contact Point Metrics.} \textbf{A)} 
 and \textbf{B)} Identification of the contact point in the pressure-volume recorded data for the hemispherical gripper ($\bar{h} = 0.052 $ and $\bar{t} = 0.048 $). The dashed line represents the reference data recorded during the deflation of the hemispherical gripper with no object. \textbf{C)} Contact metric $D_{contact}$ and the corresponding threshold value. The arrows represent the progression in time during the experiment.}
    \label{suppfig:contact}
\end{figure*}

To locate the contact point from the pressure-volume response, we introduce the metric $D_{\text{contact}}$ given by
\begin{equation}
    D_{\text{i,contact}}=\frac{P_{\text{i,norm}}^{\text{ref}}-P_{\text{i,norm}}^{\text{exp}}}{\overline{P}_{\text{norm}}^{\text{ref}}}
    \label{suppeq:contactmetric}
\end{equation}
where $P_{\text{norm}}=P/(E (h/R)^3)$ is the normalized pressure, $P_{\text{i,norm}}^{\text{exp}}$ is the normalized experimental pressure at time increment i, $P_{\text{norm}}^{\text{ref}}$ is the normalized reference pressure at time increment i, and $\overline{P}_{\text{norm}}^{\text{ref}}$ is the mean of the entire normalized reference pressure deflation curve (since it is collected prior to the live trial).
Our final evaluation utilizes a 30 point central moving mean to smooth the signal.
We apply the same threshold $D_{\text{contact}} = 0.03$ to all experimental trials, and, when exceeded, we record the onset of contact.
Fig. \ref{suppfig:contact}A shows the deflation pressure-volume curve of the electronic chip grasping trial, wherein the star represents the labeled contact point.
Fig. \ref{suppfig:contact}B depicts an inset from the deflation plot in the neighborhood of the contact point and
Fig. \ref{suppfig:contact}C highlights the evaluated $D_{\text{contact}}$ and threshold used for detecting the contact point from the pressure-volume curve.

To identify the buckling point within the pressure-volume response, we take advantage of the slope change induced by the onset of the buckling instability.
Towards this goal, we use the MATLAB Signal Processing Toolbox 'ischange' function on the slope of each point on the curve, setting a predetermined threshold to filter out noise.
We mask out the first $ 0.25 \cdot \left|\Delta V/V_0\right|$ on the $x$-axis before searching for the buckling point to avoid false flags from the contact interaction.

\begin{figure*}
    \centering
    \includegraphics[width=\textwidth,keepaspectratio]{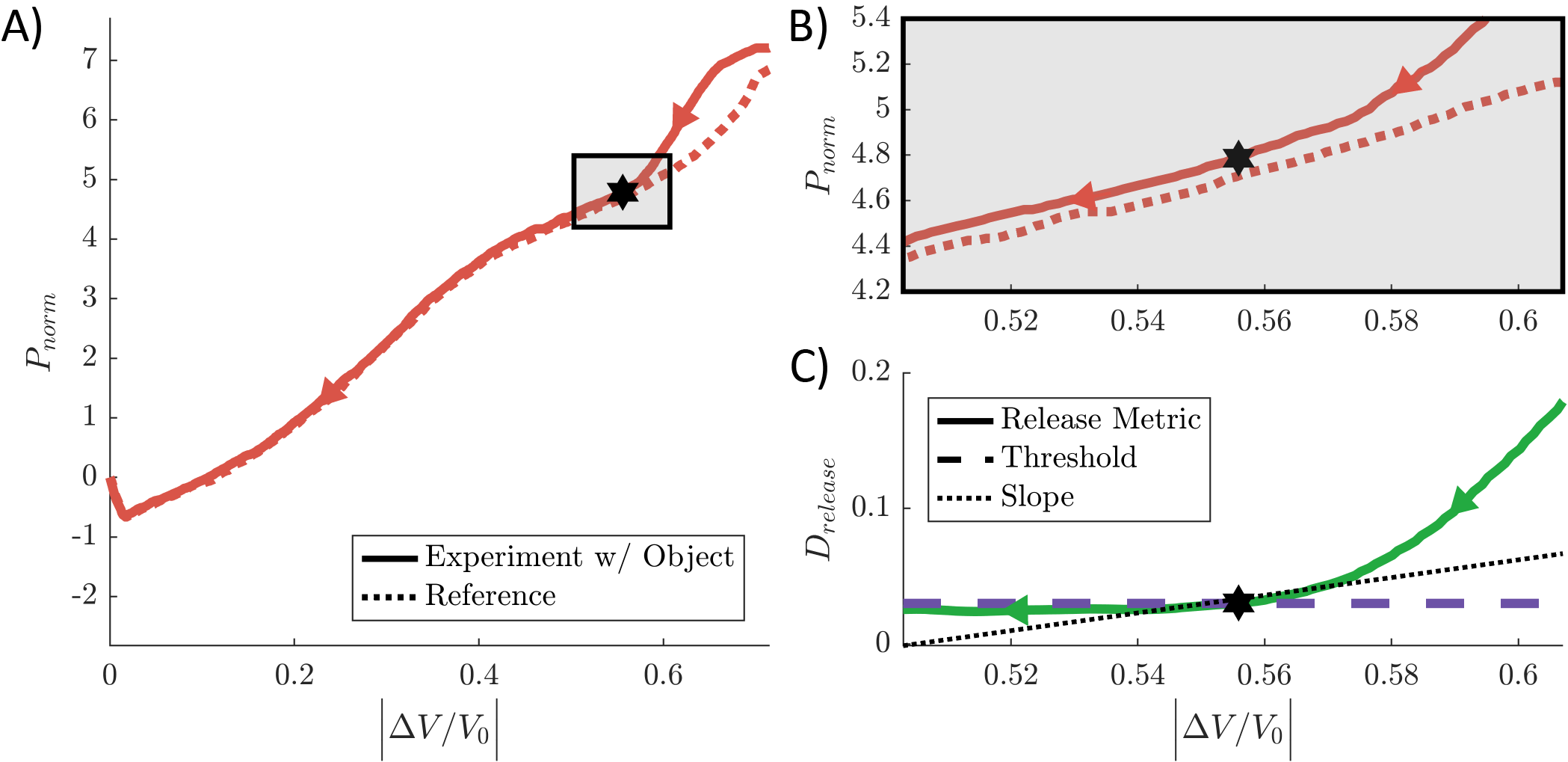}
     \caption{\textbf{Release Point Metrics.} \textbf{A)} and \textbf{B)} Identification of the release point in the pressure-volume recorded data for the hemispherical gripper ($\bar{h} = 0.052 $ and $\bar{t} = 0.048 $). The dashed line represents the reference data recorded during the deflation of the hemispherical gripper with no object. \textbf{C)} Release metric $D_{release}$ and the corresponding threshold value. The arrows represent the progression in time during the experiment.}
 
    \label{suppfig:release}
\end{figure*}

The characterization of the release point is accomplished similarly to the contact point.
Indeed, the release phase of an object begins when the pressure-volume curve of the hemispherical gripper deviates from the reference curve obtained with no object.
Then, once the object is fully removed from the hemisphere, the experimental and the reference curves overlap.
We introduce a threshold based on the normalized pressure difference and seek the first point that decreases below that preset value.
The equation for the metric we introduce is given by 
\begin{equation}
    D_{\text{i,release}}=\frac{P_{\text{i,norm}}^{\text{exp}}-P_{\text{i,norm}}^{\text{ref}}}{\overline{P}_{\text{norm}}^{\text{ref}}}
    \label{suppeq:releasemetric}
\end{equation}

where the metric is evaluated at each discrete point i, and a 30 point central moving mean is used to smooth out the signal.
To this metric, we introduce an additional step to monitor the slope of the pressure-volume curve at the point of intersection before classifying it as a release point.
Because the pressure of the hemisphere during the idle state can be similar when holding and not holding an object, the reference and experimental curves can be very close in value at the beginning of the inflation cycle.
Thus, if we apply a threshold directly, we risk to falsely flag the initial point as the release point.
As the inflation cycle progresses from the initial point, the metric increases due to the influence of the object.
The metric decreases back below the threshold after release.
To avoid false flags, we validate that the metric is decreasing by checking the slope at the threshold intersection point (which is computed using a linear polyfit of the 30 surrounding data points).
Fig. \ref{suppfig:release}A shows the inflation pressure-volume curve for the electronic chip grasping trial.
Fig. \ref{suppfig:release}B depicts an inset of the inflation plot near the release point.
Fig. \ref{suppfig:release}C highlights $D_{\text{release}}$ and the threshold used for finding the release point from the pressure-volume curve.
Fig. \ref{suppfig:release}C also demonstrates the step of finding the slope at the intersection point to ensure the metric is decreasing.

\subsubsection{C. Maximum Payload-to-Weight Ratio}

\begin{figure*}
    \centering
    \includegraphics[keepaspectratio]{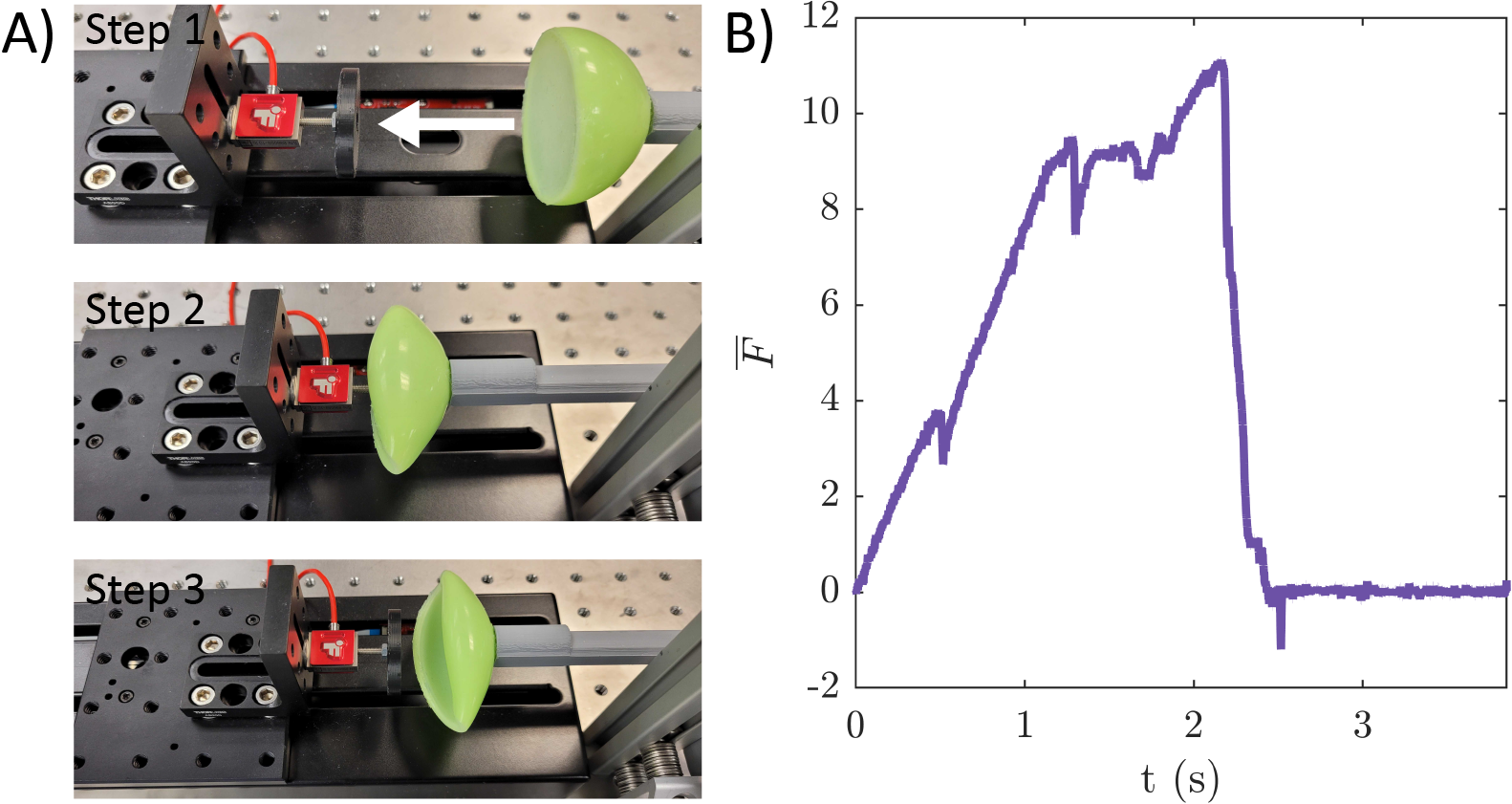}
    \caption{\textbf{Payload-to-Weight Experiment.}  \textbf{A)} Experimental snapshots of the payload-to-weight experiment. \textbf{B)} Time response of the normalized grasping force $\bar{F}$ for an hemispherical gripper ($\bar{h} = 0.052 $ and $\bar{t} = 0.048 $). }
    \label{suppfig:weightratio}
\end{figure*}

To highlight the superior properties of soft grippers based on hemispherical shells, we characterize the payload-to-weight ratio using a different experimental setup.
Here, the hemispherical gripper ($\bar{h} = 0.052 $ and $\bar{t} = 0.048 $) is mounted and aligned with a motorized linear stage (Thorlabs LTS300 Translation Stage).
A calibrated load cell (Futek LSB201) is mounted to the linear stage and equipped with a 3D-printed base disk to be grasped by the gripper.
We then connect the load cell to a Futek IAA100 Strain Gauge Analog Amplifier, which relays data through a National Instruments USB-6009 data acquisition device to a custom MATLAB script.
We follow a three step process (Fig. \ref{suppfig:weightratio}A) to collect the maximum grasping force as shown below

\begin{enumerate}
    \item[] \textbf{Step 1:} The linear stage is initialized to a starting position away from the hemispherical gripper. The gripper position is manually adjusted to put the gripper in contact with the disk attached to the load cell. 
    \item[]  \textbf{Step 2:} The gripper is deflated by 30 mL using the syringe pump and the disk is encapsulated.
    \item[] \textbf{Step 3:} The linear stage is activated at a speed of 8.0 cm/s to pull back until the grasped disk is no longer encapsulated by the gripper while the load cell is recording the reaction force. 
\end{enumerate}

Fig. \ref{suppfig:weightratio}B represents the time response of the normalized force $\overline{F}=F/(mg)$, where $m$ is the mass of the gripper ($ m = 9.6$ g) and $g$ is gravitational acceleration.
We repeat the experiment 10 times and obtain the maximum force from each time response dataset to calculate the maximum payload-to-weight ratio, which was found to be 11.70 with a standard deviation of 1.45.

\end{document}